\documentclass[pdflatex,sn-mathphys-num]{sn-jnl}

\usepackage{graphicx}%
\usepackage{multirow}%
\usepackage{amsmath,amssymb,amsfonts}%
\usepackage{amsthm}%
\usepackage{mathrsfs}%
\usepackage[title]{appendix}%
\usepackage{xcolor}%
\usepackage{textcomp}%
\usepackage{manyfoot}%
\usepackage{booktabs}%
\usepackage{algorithm}%
\usepackage{algorithmicx}%
\usepackage{algpseudocode}%
\usepackage{listings}%
\usepackage{siunitx} %

\theoremstyle{thmstyleone}%
\theoremstyle{thmstyletwo}%

\theoremstyle{thmstylethree}%

\begin{document}

\title[Article Title]{Wafer Map Defect Classification Using Autoencoder-Based Data Augmentation and Convolutional Neural Network}


\author{\fnm{Yin-Yin} \sur{Bao}}\email{bao17640667213@163.com}

\author*{\fnm{Er-Chao} \sur{Li}$^*$}\email{lecstarr@163.com}

\author{\fnm{Hong-Qiang} \sur{Yang}}\email{19894476449@163.com}

\author{\fnm{Bin-Bin} \sur{Jia}}\email{ jiabinbin@lut.edu.cn}

\affil{\orgdiv{College of Electrical Engineering and Information Engineering}, \orgname{Lanzhou University of Technology}, \orgaddress{\city{Lanzhou}, \postcode{730050}, \state{Gansu}, \country{China}}}


\abstract{In semiconductor manufacturing, wafer defect maps (WDMs) reveal critical defect patterns essential for diagnosing issues and improving process yields. However, accurate categorization of WDM defects faces significant challenges due to noisy data, unbalanced defect classes, and the complexity of failure modes. To address these challenges, this study proposes a novel method that combines self-encoder-based data enhancement with a convolutional neural network (CNN). By introducing noise in the latent space to reconstruct the WDM, the self-encoder not only enhances data diversity but also mitigates the problem of class imbalance, thereby improving the model’s generalization ability. The augmented dataset is then utilized to train the CNN, enabling it to extract hierarchical features and achieve precise classification of both common and rare defect patterns. Experiments conducted on the WM-811K dataset demonstrate that the proposed method achieves a classification accuracy of 98.56\%, outperforming Random Forest, SVM, and Logistic Regression by 19\%, 21\%, and 27\%, respectively. These results highlight the effectiveness of the proposed approach, offering a robust and accurate solution for wafer defect detection and classification.
}




\maketitle

\section{Introduction}\label{sec1}
Semiconductor manufacturing is a highly intricate process that involves numerous stages of wafer fabrication, where even minor defects can lead to significant yield losses~\cite{wei2023wafer,de2023semi,yin2019stone}. As the size of semiconductor components continues to shrink due to advances in manufacturing technology, ensuring the reliability and quality of the wafers has become increasingly essential~\cite{mills1997overview}. Yield analysis techniques, such as wafer map failure pattern recognition, play a central role in diagnosing and preventing manufacturing-related defects before they propagate through the production pipeline~\cite{yoon2022semi,ishida2019deep}. The ultimate goal of failure pattern recognition is to accurately classify different failure modes observed in wafer maps, which aids in quickly identifying root causes and making necessary corrections to the manufacturing process~\cite{ferris1989defects,poehls2021review,kim2021adversarial}.

Wafer maps are visual representations of the functional or non-functional dies (individual semiconductor components) on a semiconductor wafer~\cite{maksim2019classification,shim2023learning}. They display spatial patterns where defects develop, often revealing systematic issues related to equipment malfunction, process variation, or material inconsistencies~\cite{hansen1998use}. Recognizing these patterns is critical for improving overall yield and reducing production downtime. However, automated recognition of these failure patterns is a challenging task due to the complex and noisy nature of the wafer maps~\cite{kang2015using}. Noise in wafer map data can originate from various factors, including measurement inaccuracies, environmental effects, or random variability in the manufacturing process, all of which can obscure the true failure patterns~\cite{taha2017clustering,alawieh2017identifying}.

Traditional techniques for wafer map failure pattern recognition have mainly relied on rule-based methods, statistical analysis, and conventional machine learning approaches. Early efforts included manual inspection or heuristic algorithms that relied on handcrafted features and predefined rules to detect specific failure patterns (e.g., center, edge, or ring patterns)~\cite{geng2023mixed,nakazawa2018wafer}. Unfortunately, these methods tend to be limited in their generalization ability because they struggle to adequately capture the diversity of failure patterns present in real-world manufacturing data. Moreover, they often rely on domain-specific knowledge and require extensive fine-tuning of parameters~\cite{piao2023cnn,sumikawa2013pattern}.

As an alternative to rule-based approaches, machine learning techniques, such as Support Vector Machines (SVM)~\cite{hearst1998support}, K-Nearest Neighbors (KNN)~\cite{peterson2009k}, and decision trees~\cite{de2013decision}, have been applied to improve the classification of wafer map failure patterns. These methods focus on learning features directly from the data, rather than depending on predefined heuristics. For example, SVMs have been popular for pattern recognition due to their ability to handle high-dimensional feature spaces. However, despite offering improvements over traditional methods, machine learning classifiers are highly sensitive to noisy data, which can severely affect their performance~\cite{piao2023cnn,sumikawa2013pattern}. Additionally, they rely on feature engineering, which requires significant human effort and domain expertise, and may not be effective in dealing with complex, nuanced failure patterns~\cite{wu2014wafer}.

In more recent years, deep learning, particularly Convolutional Neural Networks (CNNs), has brought substantial advances in the field of image recognition and has been successfully applied to wafer map pattern classification tasks~\cite{yin2024evaluation}. CNNs automatically learn hierarchical representations of data and have achieved excellent results in various computer vision domains~\cite{yin2024ipev}. For wafer maps, CNNs have been used to classify failure patterns more efficiently than previous machine learning methods due to their robustness in learning spatial features. For instance, several researchers have shown that CNNs outperform conventional techniques in wafer map classification by extracting both low-level and high-level features through their layered structure~\cite{wang2020defect}. However, a major limitation remains: CNNs are still susceptible to noise in the data. Wafer maps, in particular, often contain noisy artifacts, such as random defects that can mislead the CNN model toward inaccurate classification~\cite{phua2020semiconductor}.

To address this issue, recent efforts have explored the use of denoising methodologies in the context of deep learning. Denoising Convolutional Neural Networks (DCNNs), for example, have been developed to remove noise from input data, thereby allowing the network to focus on the relevant and informative features~\cite{yu2019stacked}. Autoencoders, generative models, and denoising filters have also seen applications in various image processing tasks, where reducing noise significantly boosts overall performance~\cite{liou2014autoencoder}. These techniques have demonstrated their advantages in denoising classic image problems such as denoising grayscale or color images, text recovery, and image inpainting~\cite{zhai2018autoencoder}.

Some studies in wafer map failure analysis have started experimenting with noise-reduction approaches integrated into CNNs~\cite{cheon2019convolutional}. For instance, autoencoder architectures have been used to preprocess wafer map images by cleaning up noisy pixels before feeding them to a classification network. Noise-aware models have been shown to improve classification accuracy when working with noisy datasets by learning to distinguish useful patterns from irrelevant noise~\cite{kim2023mixed}. These studies show that noise reduction methods can be useful, but more research is still needed to fully use denoising in the field of wafer map failure pattern recognition. Moreover, a significant issue lies in the insufficiency of training data of wafer maps with well-annotated. The lack of well-annotated public datasets hampers the performance of the model and restricts the development of related algorithms~\cite{bhatnagar2022semiconductor,ferris1989defects,kim2021adversarial}.

Building on the advancements in deep learning, we propose a novel architecture for wafer map failure pattern recognition based on data augmentation through autoencoder models and Convolutional Neural Networks. Our approach leverages the data augmentation capabilities of autoencoder models alongside convolutional neural networks, significantly enhancing the model’s capacity to detect and classify failure patterns even in the presence of substantial noise. Unlike traditional methods constrained by limited sample sizes, our integrated architecture is capable of learning from larger datasets, combined with the power of convolutional neural networks, leading to more robust and accurate wafer map pattern recognition.Our key contributions are as follows:

\begin{enumerate}
    \item \textit{Preparation of the wafer map dataset.} We  primarily focuses on classifying eight types of defective wafer maps (Center, Donut, Edge-loc, Edge-ring, Local, Near-full, Random, and Scratch) from the actual database making a total of 8 wafer map categories. To enhance the model's performance, data augmentation was employed to balance the imbalanced data classes.
    
    \item \textit{Data augmentation is used to imporve size of the dataset.} Data augmentation is implemented through the use of an autoencoder model for for dimensionality reduction into the latent space, where noise is introduced, and the wafet map data is reconstructed, ultimately increasing the sample size and improving the model's performance.
    
    \item  \textit{We propose  a deep convolutional neural network (CNN) that is trained on an augmented dataset, leveraging data augmentation techniques to enhance the diversity and robustness of the training data.} This augmented dataset allows the network to generalize better to unseen examples, improving its overall performance. In addition to designing and training the CNN,as shown in Fig~\ref{fig:fig1}. we will compare its performance with traditional tools and algorithms that are commonly used for the same task. This comparison will be based on several key metrics, such as accuracy, precision, recall, and computational efficiency, to provide a comprehensive evaluation of the advantages and limitations of our deep learning approach relative to established methods.
    
    \item \textit{We conducted extensive experiments to demonstrate that the proposed method achieves optimal performance across various benchmarks.} These experiments were designed to thoroughly evaluate the robustness, accuracy, and scalability of our approach under different conditions. In addition, we performed ablation studies to systematically validate the effectiveness of each component of our model. By selectively removing or altering specific elements of the architecture, we were able to identify the contributions of each part to the overall performance, further confirming the superiority of the current method.

\end{enumerate}

In summary, this paper presents a novel and effective approach for wafer map failure pattern recognition by combining the strengths of autoencoder-based data augmentation and deep convolutional neural networks. Our method addresses the challenges posed by imbalanced datasets and noise, enabling the model to learn more robustly from a broader and more diverse set of samples. We demonstrate through comprehensive experiments, including ablation studies, that our architecture significantly outperforms traditional methods, offering improved accuracy, precision, and generalization capabilities. The proposed solution holds promise for advancing the state of the art in wafer map classification, providing a reliable and scalable tool for defect detection in semiconductor manufacturing.
\section{Dataset and Evaluation Protocol}
\subsection{Dataset}
\begin{figure*}
    \centering
    \includegraphics[width=0.98\textwidth]{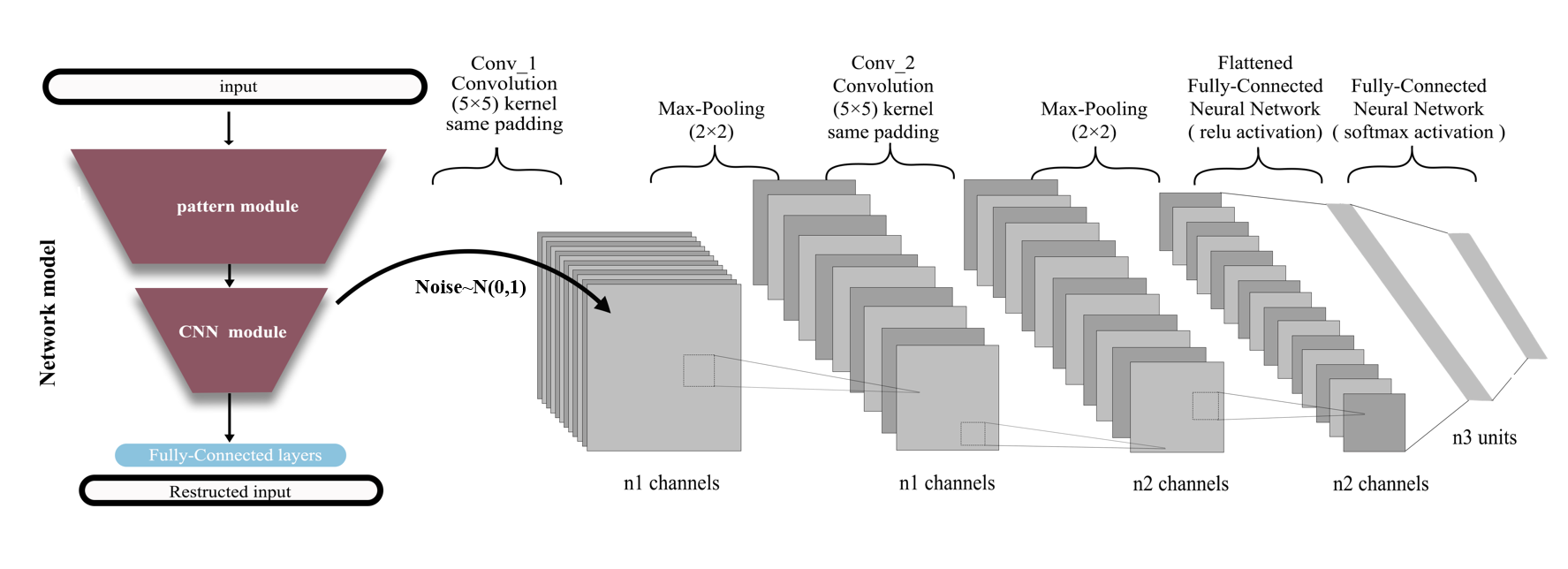}
    \caption{The architecture of the proposed network model consists of two main modules: the pattern module and the CNN module. The pattern module extracts initial spatial features from the input, which are further processed by the CNN module consisting of two convolutional layers with 5x5 kernels and same padding, followed by max-pooling layers (2x2). Gaussian noise sampled from N(0,1) is introduced to enhance robustness. The processed features are then flattened and passed through fully connected neural network layers with ReLU activation, followed by a final fully connected layer with softmax activation to produce the final class predictions (n3 units). The entire architecture is designed to capture and reconstruct key patterns from the input data.}
    \label{fig:fig1}
\end{figure*}

To demonstrate the effectiveness of the proposed model algorithm, we utilized the WM811K semiconductor dataset for our experiments~\cite{bhatnagar2022semiconductor}. This comprehensive wafer dataset contains 811,457 wafer images along with supplementary information such as wafer core dimensions, batch numbers, and wafer indices. The dataset was collected from 47,543 physical lots from a FAB, with each lot consisting of 25 wafers. However, although 47,543 lots would yield  1,557,325 wafers, the dataset contains only 811,457 wafer images. 

\begin{figure*}
    \centering
    \includegraphics[width=0.98\textwidth]{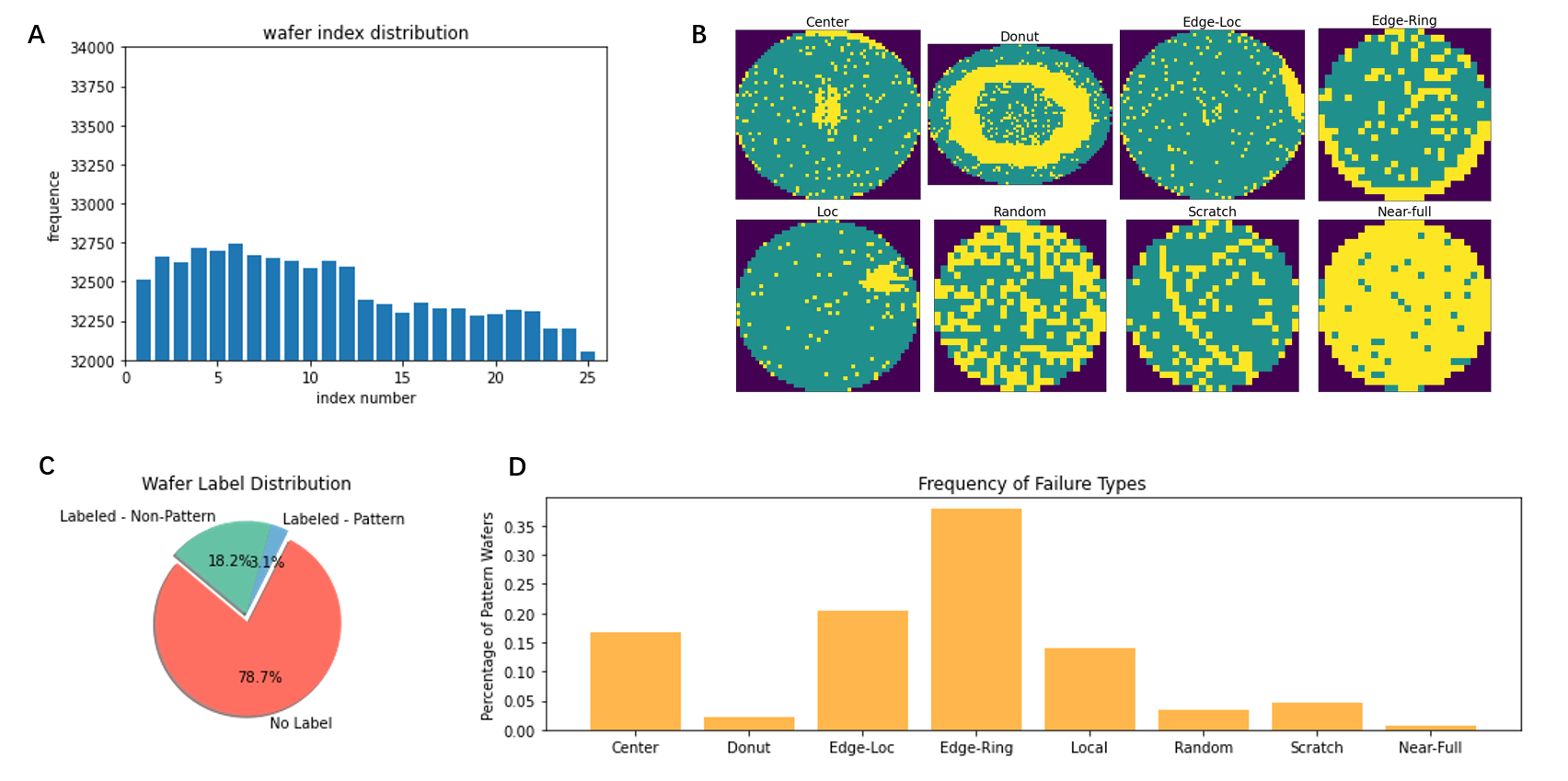}
    \caption{Visual Analysis of Wafer Defect Data. (A) Distribution of wafer indices, indicating the frequency of wafers across different index ranges. (B) Sample images of different wafer failure types, including Center, Donut, Edge-Loc, Edge-Ring, Loc (Localized), Random, Scratch, and Near-Full. (C) Distribution of wafer labels, categorizing wafers as "No Label," "Labeled - Non-Pattern," and "Labeled - Pattern" with respective proportions. (D) The frequency of different failure types as a percentage of all pattern-labeled wafers, highlighting the prevalence of various defects.}
    \label{fig:fig2}
\end{figure*}

The manually labeled portion of the dataset comprises 172,950 images with 8 distinct labels (0-7), as shown in Fig~\ref{fig:fig2}A. We observed that not all lots contain 25 wafer images, which may be due to sensor failures or other unknown issues. As a result, 172,950 wafers were labeled, while the remaining 78.7\% of the wafers were unlabeled. 

As shown in Fig~\ref{fig:fig2}B and  Fig~\ref{fig:fig2}C,among the labeled wafers, 25,519 wafers (3.1\%) were defective, while the remaining 147,431 wafers were intact. This highlights the scarcity of defective samples available for training the model. As shown in Fig~\ref{fig:fig2}D.The distribution of defects is as follows: {Center: 4294, Donut: 555, Edge-Loc: 5189, Edge-Ring: 9680, Loc: 3593, Random: 866, Scratch: 1193, Near-full: 149}. Label 8 corresponds to defect-free, normal wafers, which account for 18.2\% of the dataset, while labels 0-7 represent various defective wafer patterns.

\subsection{Evaluation Metrics}
In this paper, we evaluate the experimental results using four widely recognized metrics: Precision~\cite{kane1996precision}, Recall, F1 score, and Accuracy. These metrics are especially useful for assessing recognition performance in the context of imbalanced datasets. The calculation formulas for these metrics are as follows:
\begin{align}
\text{Precision} &= \frac{TP}{TP + FP}, \\
\text{Recall} &= \frac{TP}{TP + FN}, \\
\text{F1-score} &= \frac{2 \cdot \text{Precision} \cdot \text{Recall}}{\text{Precision} + \text{Recall}}, \\
\text{Accuracy} &= \frac{TP + TN}{TP + FN + TN + FP},
\end{align}
Where TP, FN, TN, and FP denote true positives, false negatives, true negatives, and false positives, respectively.

For multi-class classification tasks, we also use two additional metrics: Mean Area Under the Curve (Mean AUC) and Mean Average Precision (Mean AP). These metrics evaluate model performance across all classes by averaging the results obtained for each class. The corresponding formulas are as follows:
\begin{align}
\text{Mean AUC} &= \frac{1}{n_\text{classes}} \sum_{i=1}^{n_\text{classes}} \text{AUC}_i, \\
\text{Mean AP} &= \frac{1}{n_\text{classes}} \sum_{i=1}^{n_\text{classes}} \text{AP}_i,
\end{align}
Where \( \text{AUC}_i \) and \( \text{AP}_i \) represent the Area Under the ROC Curve and Average Precision for the \(i^{th}\) class, respectively. \( n_\text{classes} \) denotes the total number of classes.
After training, the model is also evaluated using accuracy and AUC (Area Under the Curve) for classification. The AUC is particularly useful for measuring performance in imbalanced datasets.

The ROC-AUC score is computed by:
\begin{align}
\text{AUC} = \int_{0}^{1} TPR(FPR) \, d(\text{FPR})
\end{align}
Where:
 \( TPR \) (True Positive Rate) and \( FPR \) (False Positive Rate) are functions of the decision threshold for classification.

\subsection{Implementation Details}
\hspace*{1.5em}The experiments outlined in this study were performed on a system running the Ubuntu operating system, specifically using Python version 3.8~\cite{van1995python} as the primary programming language. The machine learning models were implemented and trained using the TensorFlow 2.4 framework ~\cite{pang2020deep}and Pytorch 1.9~\cite{imambi2021pytorch}, which provided the necessary tools for building and optimizing deep learning architectures. In terms of hardware, the system was equipped with the NVIDIA V100 GPU, a high-performance graphics card designed for accelerating deep learning computations, particularly useful for large-scale neural network training. Additionally, the used system was powered by an Intel Xeon CPU, known for its robust multi-core processing capabilities, which assisted in efficiently handling data preprocessing.This combination of software and hardware ensured that the experiments could be conducted both efficiently and at scale.

\section{Proposed methods}

\subsection{Autoencoder for Reconstruction and Latent Space Representation}

\hspace*{1.5em}The autoencoder's role in our model is to learn a compressed representation of the wafer maps in the latent space and then reconstruct the original wafer maps from this compressed space. This can be expressed in two stages: encoding and decoding.

\begin{itemize}
 \item Encoding:
Let \( X \in \mathbb{R}^{26 \times 26 \times 3} \) represent the input wafer map. The encoder compresses this input into a lower-dimensional latent representation \( Z \in \mathbb{R}^{13 \times 13 \times 64} \).

The encoding process is represented mathematically as:
\begin{align}
Z = f_{\text{encoder}}(X) = \sigma(W_e \ast X + b_e).
\end{align}

Where:
 \( W_e \) and \( b_e \) are the weights and biases of the encoder's convolutional layers.
 \( \ast \) denotes the convolution operation.
 \( \sigma \) is the ReLU activation function.

\item  Decoding:
The decoder reconstructs the wafer map from the latent space \( Z \) back to its original form \( \hat{X} \in \mathbb{R}^{26 \times 26 \times 3} \). The decoding process can be written as:
\begin{align}
\hat{X} = f_{\text{decoder}}(Z) = \sigma(W_d \ast Z + b_d).
\end{align}

Where:
 \( W_d \) and \( b_d \) are the weights and biases of the decoder's transposed convolution layers.

\hspace*{1.5em}The autoencoder is trained by minimizing the reconstruction loss: the mean squared error (MSE) between the original wafer map \( X \) and the reconstructed wafer map \( \hat{X} \):
\begin{align}
\mathcal{L}_{\text{MSE}} = \frac{1}{n} \sum_{i=1}^{n} (X_i - \hat{X}_i)^2.
\end{align}
\end{itemize}

\subsection{Adding Noise for Data Augmentation}

\hspace*{1.5em}After training the autoencoder, we add Gaussian noise to the latent space to create new wafer maps. The purpose of this is to generate synthetic data that can be used to train the CNN and improve generalization.

Let the latent representation \( Z \) of the wafer map be:
\begin{align}
Z = f_{\text{encoder}}(X).
\end{align}

To augment the data, we add noise sampled from a normal distribution with mean \( 0 \) and standard deviation \( \sigma \):
\begin{align}
Z_{\text{noisy}} = Z + \epsilon.
\end{align}
Where \( \epsilon \sim \mathcal{N}(0, \sigma^2) \).

The noised latent representation \( Z_{\text{noisy}} \) is then decoded back into the image space:
\begin{align}
\hat{X}_{\text{noisy}} = f_{\text{decoder}}(Z_{\text{noisy}}).
\end{align}

This produces a new wafer map \( \hat{X}_{\text{noisy}} \) that is similar to the original but slightly different due to the added noise.
\subsection{Data Augmentation for All Faulty Classes}

\hspace*{1.5em}The augmented data is generated for each faulty class by repeating the process of adding noise to the latent space of the corresponding wafer maps. This augmentation increases the number of samples for each class, ensuring that the model has enough training data for all defect types.

For each faulty class \( c \in \{0, 1, \ldots, 7\} \), which is `Center', `Donut', `Edge-Loc', `Edge-Ring', `Loc', `Near-full',  `Random', and
`Scratch', respectively. We generate additional wafer maps \( \hat{X}_c \) using the following steps:

\begin{align}
Z_c &= f_{\text{encoder}}(X_c) \quad \text{(for all samples of class } c),
\end{align}
\begin{align}
Z_{c, \text{noisy}} &= Z_c + \epsilon,
\end{align}
\begin{align}
\hat{X}_{c, \text{noisy}} &= f_{\text{decoder}}(Z_{c, \text{noisy}}).
\end{align}

The newly generated wafer maps \( \hat{X}_{c, \text{noisy}} \) are then added to the dataset along with their corresponding labels.

\subsection{CNN Classification}

\hspace*{1.5em}Once the dataset has been augmented, we use a Convolutional Neural Network (CNN) to classify the wafer maps into different defect types.
The CNN takes an input wafer map \( \hat{X}_{c, \text{noisy}}  \in \mathbb{R}^{26 \times 26 \times 3} \) and processes it through multiple convolutional layers to extract spatial features.

The input to the CNN is a wafer map \(  \hat{X}_{c, \text{noisy}}  \in \mathbb{R}^{26 \times 26 \times 3} \), which represents a 26x26 image with 3 color channels.
\begin{align}
\hat{X}_{c, \text{noisy}}  = \{ X_{ijk} \mid i,j = 1 \dots 26,  \ k = 1 \dots 3 \}.
\end{align}
\hspace*{1.5em}The first convolutional layer applies 16 filters (or kernels) of size \(3 \times 3\) with padding set to `same', which ensures the output size remains \(26 \times 26\).

The convolution operation can be written as:
\begin{align}
H^{(1)}_{ijk} = \sigma \left( \sum W^{(1)}_{pqkm} X_{(i+p)(j+q)m} + b_k^{(1)} \right).
\end{align}
Where:
 \( W^{(1)}_{pqkm} \) are the weights (filters) of the convolutional layer.
\( b_k^{(1)} \) is the bias for filter \(k\).
 \( X \) is the input image.
 \( \sigma(x) = \text{ReLU}(x) = \max(0, x) \) is the activation function (ReLU).

Since padding is set to `same', the size of the output feature map remains \(26 \times 26\) and the final output from this convolutional layer is:
\begin{align}
H^{(f)} \in \mathbb{R}^{26 \times 26 \times 128}
\end{align}
\hspace*{1.5em}The next step is to flatten the output of the last convolutional layer into a 1D vector so that it can be passed into fully connected layers. The flattening operation transforms the 3D tensor \( H^{(3)} \) of shape \(26 \times 26 \times 128\) into a 1D vector of size 86,528.
\begin{align}
F = \text{Flatten}(H^{(3)}) \quad \text{where} \ F \in \mathbb{R}^{86528}.
\end{align}
\hspace*{1.5em}After flattening, the network uses fully connected layers, where each neuron is connected to all neurons in the previous layer.
The first dense layer has 512 neurons, and each neuron applies a weighted sum followed by the ReLU activation function:
\begin{align}
D_1 = \sigma(W_d^{(1)} F + b_d^{(1)}).
\end{align}
Where:
 \( W_d^{(1)} \in \mathbb{R}^{512 \times 86528} \) are the weights of the first dense layer.
 \( b_d^{(1)} \in \mathbb{R}^{512} \) is the bias term.
 \( D_1 \in \mathbb{R}^{512} \) is the output of the first dense layer.
The final output layer consists of 8 neurons (corresponding to the 8 classes), and each neuron outputs the probability that the input belongs to a specific class. The softmax activation is applied to convert the raw outputs into probabilities:
\begin{align}
P(y=k|X) = \frac{e^{z_k}}{\sum_{j=1}^{8} e^{z_j}}.
\end{align}
Where:
 \( z_k = W_d^{(3)} D_2 + b_d^{(3)} \) is the raw score for class \(k\) (logit).
 \( W_d^{(3)} \in \mathbb{R}^{8 \times 128} \) are the weights of the output layer.
 \( b_d^{(3)} \in \mathbb{R}^{8} \) is the bias term.
 \( P(y=k|X) \) is the predicted probability for class \(k\).
The output of this layer is a probability vector:
\begin{align}
\hat{y} = [P(y=1|X), P(y=2|X), \dots, P(y=8|X)].
\end{align}
\hspace*{1.5em}The network is trained to minimize the categorical cross-entropy loss, which measures the difference between the true label distribution and the predicted probabilities:
\begin{align}
\mathcal{L}_{\text{cross-entropy}} = - \sum_{i=1}^{n} \sum_{k=1}^{8} y_{i,k} \log P(y_i = k | X_i).
\end{align}
Where:
\( y_{i,k} \) is a binary indicator (0 or 1) if class \(k\) is the correct label for sample \(i\).
 \( P(y_i = k | X_i) \) is the predicted probability for class \(k\).

\section{Results and Discussions}

\subsection{Data Augmentation Using Interpolation and Autoencoder Techniques}
We designed five models for comparison, including CNN, VotingClassifier, Logistic Regression, SVM, and Random Forest. The latter four models use feature extraction methods, while the CNN model employs data augmentation based on an autoencoder. The Voting Classifier model, which combines Logistic Regression, SVM, and Random Forest using a soft voting approach, employs a comprehensive set of 59 features derived from three main approaches: density-based, transformation-based, and size-based features. The density-based features consist of 13 specific attributes extracted from distinct regions of the wafer map. These regions are strategically divided into 9 central areas and 4 peripheral (or edge) areas, allowing the model to capture the density variations both at the core and along the edges of the wafer. This segmentation helps in identifying localized patterns and edge-specific anomalies that might be characteristic of certain types of defects.

\begin{figure*}
    \centering
    \includegraphics[width=0.98\textwidth]{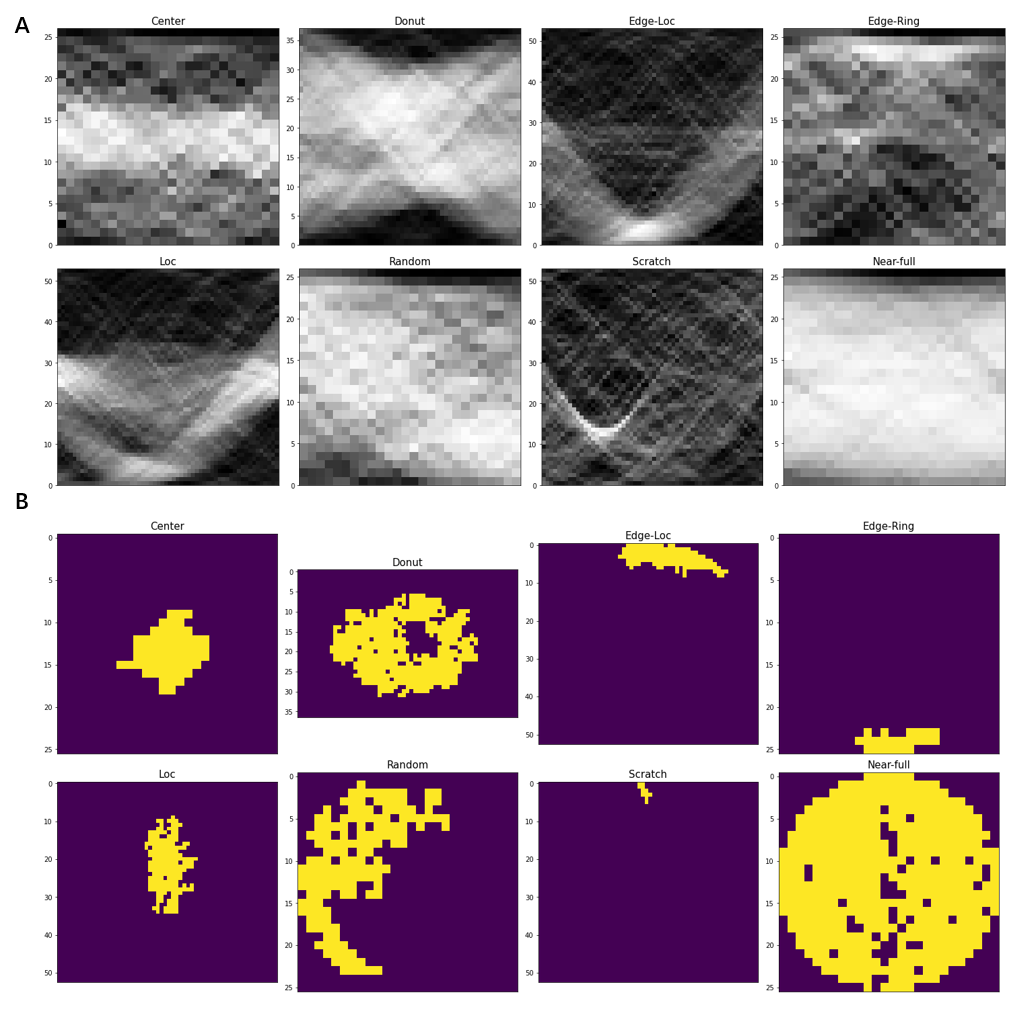}
    \caption{Visual Analysis of Wafer Defect Data. (A) Radon transform results of eight different wafer defect types, showing the transformed projections for each type, including Center, Donut, Edge-Loc, Edge-Ring, Loc (Localized), Random, Scratch, and Near-Full. The Radon transform captures the structural characteristics of defects in different directions. (B) Binary mask representations of the same eight defect types, highlighting the most prominent connected defect regions in each wafer map. The yellow regions indicate the detected defect areas after noise filtering, providing a clear view of each defect’s geometric characteristics.}
    \label{fig:fig3}
\end{figure*}

As shown in Fig~\ref{fig:fig3}A, the transformation-based features are formed by analyzing the wafer maps using Radon transformation across a range of angles from 0 to 180 degrees. The Radon transform essentially projects the wafer images at each angle, capturing the directional variations of defect patterns. After generating these projections, we utilize cubic interpolation to standardize the length of these features to 20 points each for both the mean and standard deviation of each transformed line. This results in a total of 40 transformation-based features, providing a fixed-length representation of the structural properties of defects, regardless of the original image size.

As shown in Fig~\ref{fig:fig3}B, the size-based features are derived by examining the geometric properties of the largest defect region detected in each wafer image. A total of 6 geometric attributes are extracted, including the area, perimeter, lengths of the major and minor axes, eccentricity (a measure of the region’s elongation), and solidity (which indicates the compactness of the region). These attributes help in quantifying the physical dimensions and shape characteristics of the detected defects.

Combining all these, the model leverages 59 features that collectively capture the density distribution, directional projections, and geometric properties of defects, enabling a robust analysis and classification of wafer anomalies.

\begin{figure}
    \centering
    \includegraphics[width=0.8\textwidth]{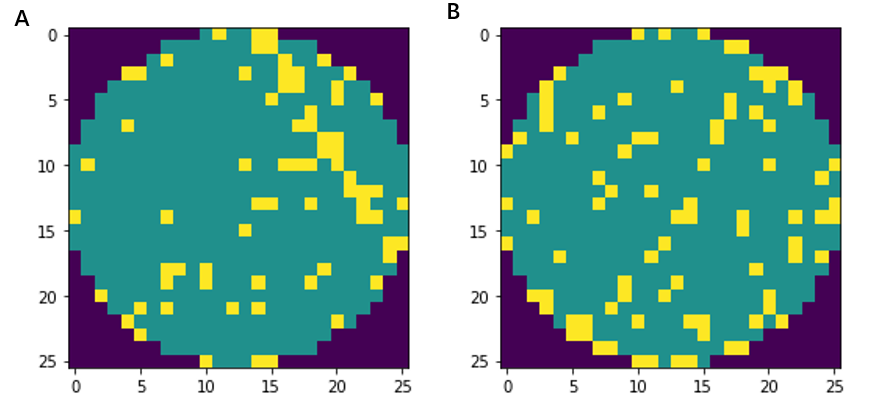}
\caption{Comparison of original and reconstructed images using a noised latent vector. (A) The original image. (B) The reconstructed image after adding noise to the latent representation. The visual differences highlight the autoencoder's ability to maintain features despite perturbations.}
    \label{fig:reconstructed_comparison}
\end{figure}

As shown in Fig~\ref{fig:fig1},we developed an autoencoder-based approach to explore wafer map image reconstruction and evaluate the impact of noise on encoded representations. The autoencoder model was specifically designed to compress image data into a lower-dimensional latent space and then reconstruct the original image from this compressed representation. To achieve this, we employed a convolutional neural network (CNN) architecture for both the encoder and decoder components of the autoencoder.

The encoder extracts key spatial features from the input image by applying convolutional operations and dimensionality reduction techniques. This process yields a compressed latent representation that captures essential information about the image while discarding less relevant details. To test the resilience of this representation, we introduced Gaussian noise into the latent space and analyzed how this perturbation influenced the reconstructed output. The addition of noise allows us to assess the robustness of the autoencoder in retaining critical visual features despite distortions.

The decoder reconstructs the image from the noised latent representation by applying a series of inverse operations, including transposed convolutions and upsampling. This inverse process aims to restore the original spatial dimensions and reconstruct the image as accurately as possible. The use of transposed convolutions enables the model to learn how to fill in missing details and enhance the output quality.

To visualize the effectiveness of our approach, we generated and compared two sets of images: one representing the original input image and the other showing the reconstructed image generated from the noised latent vector. Fig \ref{fig:reconstructed_comparison} illustrates these results, where Subfigure (A) depicts the original image and Subfigure (B) displays the reconstructed version. By comparing these two images, we can observe the model’s ability to preserve structural details and visual consistency, when subjected to noise in the encoded representation.

This experimental setup highlights the potential of using autoencoders not only for image compression and reconstruction but also for assessing the resilience of learned representations under varying conditions. Our findings suggest that the autoencoder architecture can effectively maintain essential features despite perturbations, offering promising applications in scenarios where data integrity may be compromised or altered.

\begin{figure}
    \centering
    \includegraphics[width=0.8\textwidth]{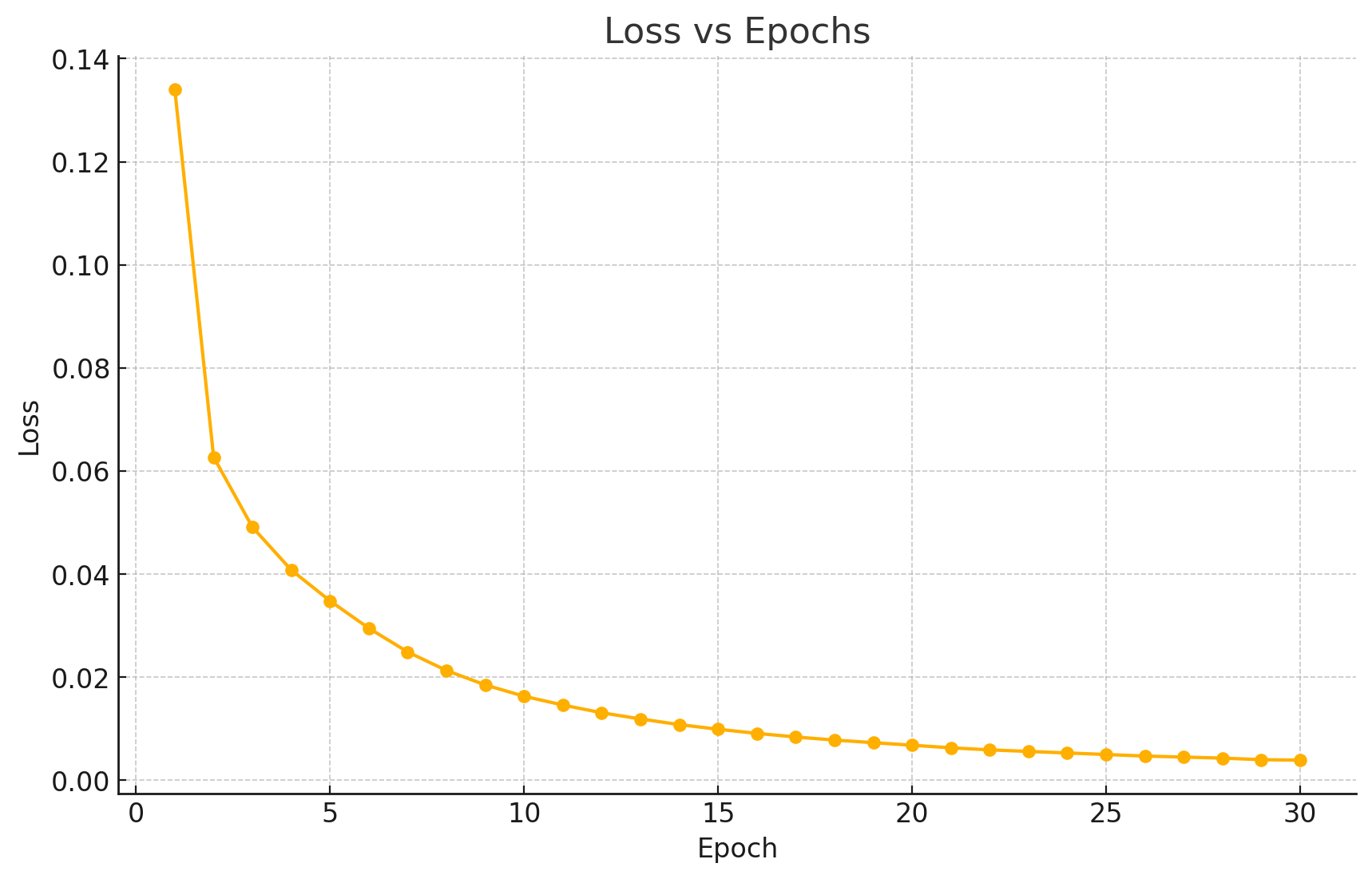}
    \caption{Training progress of the autoencoder model based on reconstruction loss. The plot depicts the reduction in reconstruction loss over 30 epochs, indicating improved model performance in reconstructing input images. The gradual decrease in loss reflects the model's increasing capability to capture and encode essential features from the input data.}
    \label{fig:reconstruction_loss}
\end{figure}

As shown in Fig~\ref{fig:reconstructed_comparison}, Fig~\ref{fig:reconstruction_loss} and Tabel~\ref{tab:augmented_data_distribution}, the graph illustrates the training progression of an autoencoder model by depicting the reduction in reconstruction loss over 30 epochs. As the training continues, the decreasing loss indicates that the model is progressively learning to encode and reconstruct the input data more accurately. This trend demonstrates the autoencoder’s ability to capture critical features and minimize discrepancies between the original and reconstructed outputs, thereby refining its performance with each epoch.
Table ~\ref{tab:target_statistics} presents the original dataset's distribution across eight classes, highlighting significant class imbalance. The dataset contains a total of 24,519 samples, unevenly distributed across classes. For instance, Class 3 (Edge-Ring) comprises 9,680 samples, while Class 7 (Scratch) has only 149 samples. This disparity is further reflected in the breakdown between training and testing sets, where classes with fewer samples have proportionally limited data for model training and evaluation.
To mitigate the impact of class imbalance, an augmentation process was applied using an autoencoder, as summarized in Table~\ref{tab:augmented_data_distribution}. This augmentation increased each class to 10,000 samples, ensuring equal representation for all classes in the dataset. The augmentation process aimed to standardize the data distribution, thereby enhancing model robustness and minimizing potential biases toward classes with initially higher sample counts. This approach provides a more balanced training dataset, contributing to improved model performance and fairness in classification tasks.
\begin{table}[t]
    \centering
    \scriptsize
        \caption{Comparison of target distribution, training statistics, and testing statistics for each class.}
    \setlength{\tabcolsep}{1pt} 
    \begin{tabular*}{0.8\textwidth}{@{\extracolsep{\fill}}cccc@{}}
        \toprule
        Class & Dataset Distribution & Training set & Testing set \\
        \midrule
        0 & \num{4,294} & \num{3,238} & \num{1,056} \\
        1 & \num{555} & \num{404} & \num{151} \\
        2 & \num{5,189} & \num{3,860} & \num{1,329} \\
        3 & \num{9,680} & \num{7,299} & \num{2,381} \\
        4 & \num{3,593} & \num{2,677} & \num{916} \\
        5 & \num{866} & \num{640} & \num{226} \\
        6 & \num{1,193} & \num{905} & \num{288} \\
        7 & \num{149} & \num{116} & \num{33} \\
        \bottomrule
    \end{tabular*}

    \label{tab:target_statistics}
\end{table}
\begin{table}[t]
    \centering
    \scriptsize
    \caption{Data distribution of each class after augmentation to reach 10,000 samples per class.}
    \setlength{\tabcolsep}{2pt} 
    \begin{tabular*}{0.4\textwidth}{@{\extracolsep{\fill}}ccc@{}}
        \toprule
        Label & Class Name & Count \\
        \midrule
        0 & Center & \num{10,000} \\
        1 & Donut & \num{10,000} \\
        2 & Edge-Loc & \num{10,000} \\
        3 & Edge-Ring & \num{10,000} \\
        4 & Loc & \num{10,000} \\
        5 & Near-full & \num{10,000} \\
        6 & Random & \num{10,000} \\
        7 & Scratch & \num{10,000} \\
        \bottomrule
    \end{tabular*}
    \label{tab:augmented_data_distribution}
\end{table}

\subsection{Evaluating the Impact of Data Augmentation on Model Performance}

The CNN-AUG model is a convolutional neural network architecture designed for classifying wafer defect patterns, enhanced with data augmentation techniques to improve its generalization capabilities. The architecture consists of multiple convolutional layers followed by fully connected layers, structured to effectively capture spatial features in wafer images.

The model begins with an input layer accepting images of size 26x26 with three channels. This is followed by a series of three convolutional layers, with 16, 64, and 128 filters, respectively, and a kernel size of 3x3. Each convolutional layer employs ReLU activation and padding to preserve the spatial dimensions of the input. These layers progressively learn higher-level features, enabling the network to detect complex patterns indicative of various defect types. After the convolutional layers, a flattening layer is applied to transform the 3D feature maps into a 1D vector, preparing the data for the subsequent dense layers.

The fully connected section of the CNN-AUG model comprises two dense layers with 512 and 128 units, respectively, each employing ReLU activation to introduce non-linearity and enhance feature learning. The final output layer consists of 8 units with a softmax activation function, producing probability distributions across the 8 classes of wafer defects. This architecture is optimized using the Adam optimizer and categorical cross-entropy loss function to achieve accurate multi-class classification.
\begin{table}[t]
    \centering
    \scriptsize
    \caption{Architecture of the CNN Model}
    \setlength{\tabcolsep}{1pt} 
    \begin{tabular*}{0.8\textwidth}{@{\extracolsep{\fill}}ccc@{}}
        \toprule
        Layer Type & Output Shape & Parameters \\
        \midrule
        Input & (26, 26, 3) & - \\
        Conv2D (3x3, 16 filters) & (26, 26, 16) & 448 \\
        Conv2D (3x3, 64 filters) & (26, 26, 64) & 9,280 \\
        Conv2D (3x3, 128 filters) & (26, 26, 128) & 73,856 \\
        Flatten & 86,528 & - \\
        Dense (512 units, ReLU) & 512 & 44,354,560 \\
        Dense (128 units, ReLU) & 128 & 65,664 \\
        Dense (8 units, Softmax) & 8 & 1,032 \\
        \midrule
        \textbf{Total Parameters} & - & \textbf{44,504,840} \\
        \bottomrule
    \end{tabular*}
    \label{tab:model_architecture}
\end{table}

In our experiments, we aimed to address class imbalance using an autoencoder-based data augmentation technique. As shown in Table ~\ref{tab:augmented_data_distribution}, we augmented the dataset such that each class contained 10,000 samples, ensuring balanced representation across all classes. This augmentation was crucial to overcoming the inherent skew in the original dataset distribution (Table ~\ref{tab:target_statistics}), where some classes were heavily underrepresented. To maintain consistency with our evaluation protocol, we first divided the dataset into training and testing sets using a 4:1 ratio before data augmentation. This split allowed for sufficient samples in both sets, supporting robust training and comprehensive evaluation.
Training and Validation Loss and Accuracy

\begin{figure}
    \centering
    \includegraphics[width=0.95\textwidth]{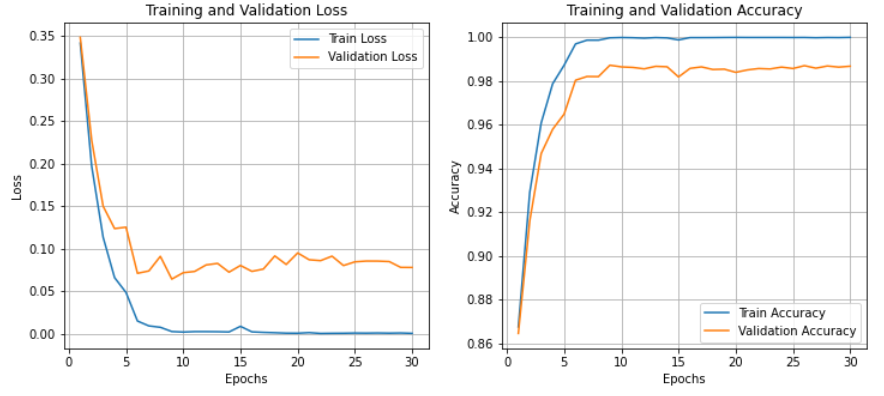}
    \caption{ Model Training and Validation Performance Before Data Augmentation
The plots depict the training and validation loss (left) and accuracy (right) of the model over 30 epochs before applying data augmentation. The model shows a noticeable gap between training and validation loss after the initial epochs, indicating potential overfitting. Similarly, the validation accuracy plateaus around 85\% while the training accuracy continues to increase, suggesting that the model struggles to generalize to the validation set due to class imbalances in the original dataset.}
    \label{fig:6}
\end{figure}

We proceeded to train our model on the original, unbalanced dataset for 30 epochs, recording its performance metrics throughout. As illustrated in Fig~\ref{fig:6}, the training loss decreased steadily over time, nearly approaching zero. However, we observed that the validation loss initially declined but began to rise after approximately 20 epochs. This divergence indicated a classic case of overfitting, where the model continued to memorize the training data while failing to generalize to new, unseen data effectively. This observation was further reinforced by the accuracy curves, which revealed that while the training accuracy consistently improved, reaching nearly 99\%, the validation accuracy plateaued around 85\%. Such a discrepancy suggests that the model struggled to maintain its performance on the validation set, likely due to the imbalance in class distributions within the original dataset.

\begin{figure}
    \centering
    \includegraphics[width=0.95\textwidth]{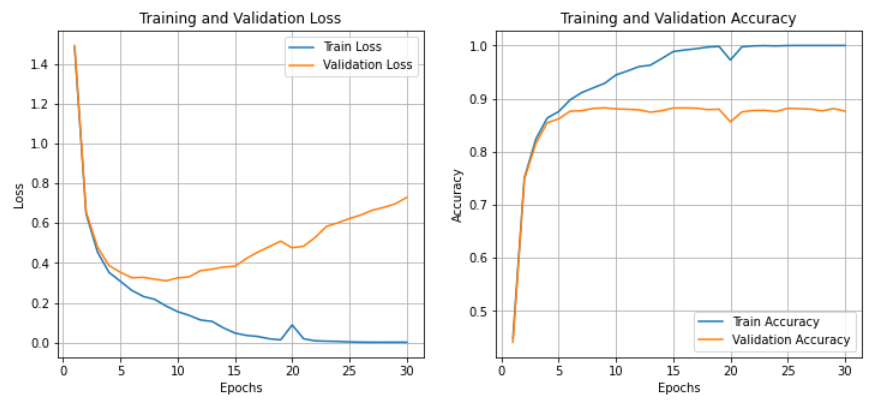}
    \caption{Model Training and Validation Performance After Data Augmentation The plots show the training and validation loss (left) and accuracy (right) of the model over 30 epochs after applying data augmentation using the autoencoder. The training and validation losses exhibit a more consistent trend with reduced divergence, highlighting the impact of balanced class representation. Additionally, the validation accuracy is significantly improved, reaching approximately 98\%, which indicates that the model achieves better generalization and overall stability after augmentation.}
    \label{fig:7}
\end{figure}

\begin{figure*}
    \centering
    \includegraphics[width=0.9\textwidth]{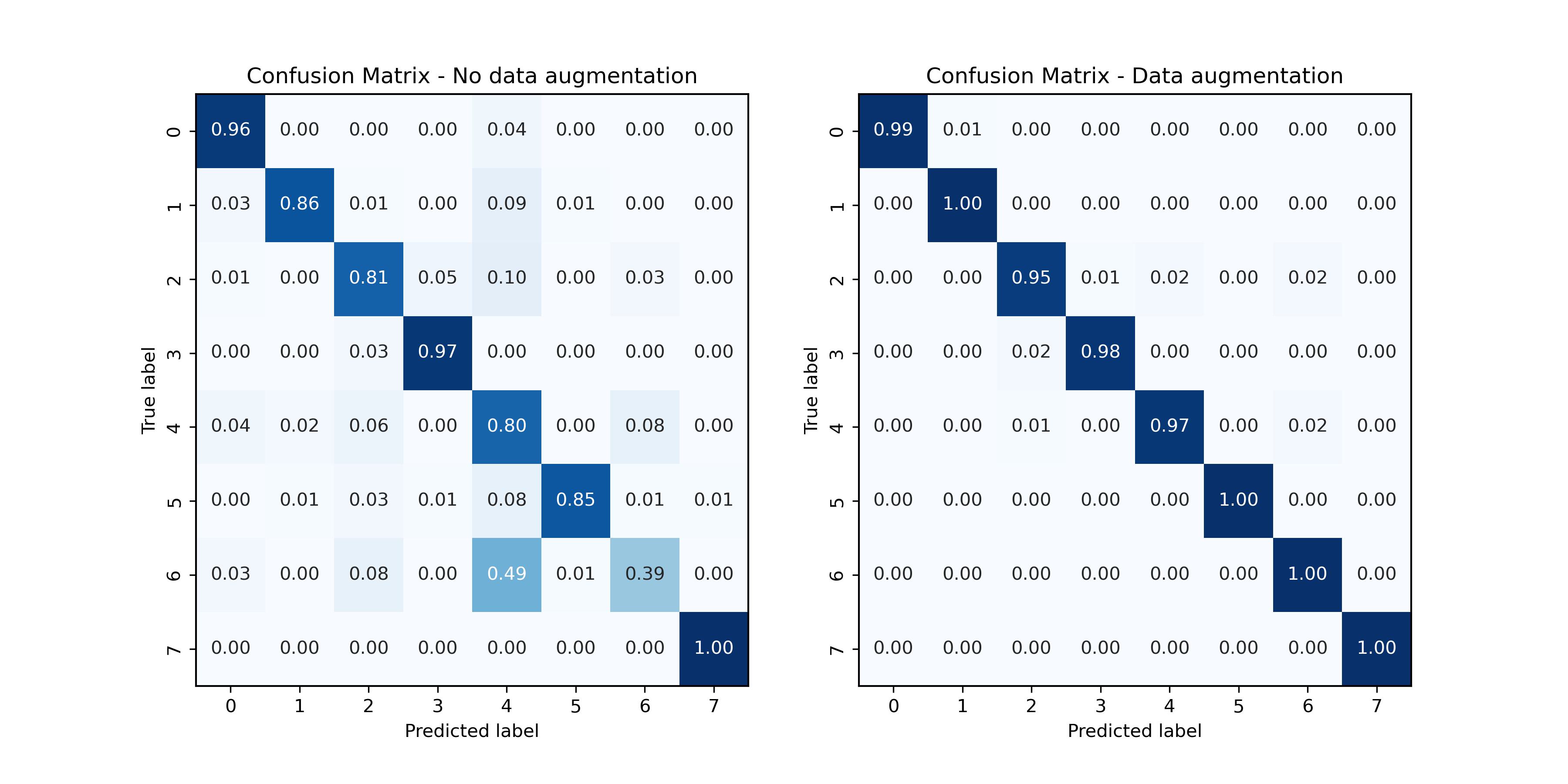}
    \caption{The confusion matrix on the left illustrates the classification performance of the model without and data augmentation. Diagonal values represent the model's accuracy for each class, while off-diagonal values indicate misclassifications. The results show that while the model performs well in recognizing certain classes such as Class 3 (Edge-Ring) and Class 0 (Center), it struggles with others, such as Class 4 (Loc) and Class 6 (Random). The overall lower accuracy for these classes reflects the class imbalance and insufficient representation in the original dataset.The confusion matrix on the right displays the model's performance after applying data augmentation. A clear improvement is evident, with higher diagonal values indicating better classification accuracy across all classes. The model achieves near-perfect classification for most classes, notably improving its performance on previously underrepresented classes such as Class 4 (Loc) and Class 6 (Random). The uniformity in predictions after augmentation highlights the effectiveness of using the autoencoder to achieve balanced class distributions and improved generalization.}
    \label{fig:fig8}
\end{figure*}

To mitigate this overfitting and class imbalance issue, we applied an autoencoder-based data augmentation strategy to balance the dataset. Following this augmentation, we retrained the model under the same conditions and observed significant changes in performance, as depicted in Fig~\ref{fig:7}. Notably, the validation loss demonstrated a more consistent decline throughout the training process, stabilizing at a value comparable to the training loss. This indicated that the model was no longer overly specialized to the training data and had improved its capacity to generalize. Additionally, the validation accuracy saw a substantial increase, reaching around 98\%, indicating a more robust and generalized learning process. The observed reduction in overfitting and the substantial improvement in validation accuracy emphasize the effectiveness of our data augmentation approach in enhancing model generalization and performance.
Confusion Matrix Analysis
To gain further insights into the impact of data augmentation on classification performance, we evaluated the model using confusion matrices, which offer a detailed perspective on how well the model distinguishes between different classes. Fig~\ref{fig:fig8} shows the confusion matrices before and after data augmentation. Before applying data augmentation, the model displayed relatively high accuracy for well-represented classes such as Center (96\%), Edge-Ring (97\%), and Near-full (85\%). However, the model’s performance was significantly lower for underrepresented classes such as Loc (80\%) and Random (39\%). The poor performance in these classes underscores the limitations of training on a dataset with pronounced class imbalances, as the model struggles to learn adequate distinguishing features for infrequent classes.
After implementing data augmentation, we observed substantial improvements in the confusion matrix, as shown in Fig ~\ref{fig:fig8}. The model achieved nearly perfect classification accuracy across all classes, with notable improvements in several key areas. For example, the model’s accuracy in recognizing Center reached 99\%, while Donut achieved 100\%, and Edge-Loc improved to 95\%. More importantly, the previously underperforming classes showed marked gains: the accuracy for Loc increased from 80\% to 97\%, and the accuracy for Random rose significantly from 39\% to 100\%. This considerable enhancement in performance for these classes demonstrates the effectiveness of our autoencoder-based data augmentation technique in creating a more balanced and representative training dataset.
In summary, our experiments demonstrated that addressing class imbalance through an autoencoder-based data augmentation strategy can significantly improve model performance. By augmenting each class to contain 10,000 samples, we achieved a balanced dataset that provided uniform representation for all classes. This approach not only reduced the model’s tendency to overfit but also resulted in a substantial increase in validation accuracy, from 85\% to approximately 98\%. Furthermore, the confusion matrix analysis revealed that the model’s classification accuracy improved considerably across all classes, particularly for those that were initially underrepresented, such as Loc and Random. These improvements underscore the importance of balanced datasets in training deep learning models and validate the efficacy of our data augmentation approach in enabling the model to learn more generalized and distinguishing features across all classes.

\subsection{A Comparative Study of Augmentation Tools for Wafer Defect Detection}
We conducted a comprehensive evaluation of various classification models, including Logistic Regression, Support Vector Machine (SVM), Random Forest, and a Voting Classifier, alongside our custom-designed Convolutional Neural Network with data augmentation (CNN-AUG). To assess the effectiveness of these models, we focused on key performance metrics such as accuracy, precision, recall, F1-score, Area Under the Curve (AUC), and average precision (AP). Given that our dataset involves an 8-class classification task, it was essential to employ a One-vs-Rest (OvR) strategy to evaluate each class individually against the remaining classes. This approach allowed us to obtain performance metrics that reflect the model's discriminative power in distinguishing each class from all others.
We conducted an extensive evaluation of various classification models, including Logistic Regression, Support Vector Machine (SVM), Random Forest, and a Voting Classifier. These traditional machine learning models were benchmarked against our custom-designed Convolutional Neural Network with data augmentation (CNN-AUG) to assess their effectiveness across key performance metrics: accuracy, precision, recall, F1-score, AUC, and average precision (AP). The results are summarized in Table \ref{tab:model_performance}.

The Logistic Regression model, known for its simplicity and interpretability, demonstrated relatively modest performance metrics. It achieved an accuracy of 0.7176, a precision of 0.5172, and a recall of 0.5147, indicating moderate ability in correctly identifying positive instances across all classes. Despite an AUC value of 0.9345 suggesting reasonable discriminative power between classes, the low AP score of 0.5653 highlighted the model's difficulty in maintaining precision, especially in a multi-class classification setting. This outcome reflects the limitations of logistic regression in handling complex patterns and class imbalances in the dataset.The SVM model showed a marked improvement over logistic regression. With an accuracy of 0.7770 and a precision of 0.7627, it demonstrated better precision in correctly classifying positive instances. The recall score of 0.7244 suggested the model's ability to capture a larger proportion of true positives. 

Additionally, the AUC score of 0.9638 indicated that the SVM could distinguish between classes with high confidence. An AP score of 0.8141 further demonstrated the model’s enhanced performance in achieving a balanced trade-off between precision and recall, which is crucial for multi-class scenarios.The Random Forest model excelled further, capitalizing on its ensemble-based approach. It achieved an accuracy of 0.8125 and a notably higher precision of 0.8482. This indicates that the model consistently classified positive instances with minimal false positives. The recall score of 0.7090, combined with an AUC of 0.9759, demonstrated its strong overall discriminative capability. Moreover, the AP score of 0.8476 indicated a high level of confidence in its predictions across multiple classes. These metrics underscore Random Forest’s ability to capture complex patterns through its decision-tree-based structure, which aggregates multiple weak learners to yield a strong classifier. The Voting Classifier,which combine above all models, exhibited robust performance across the board. With an accuracy of 0.7951, precision of 0.8050, and recall of 0.7322, it achieved a balanced outcome by integrating the predictions of Logistic Regression, SVM, and Random Forest. This balanced approach was further reflected in its AUC of 0.9709 and AP score of 0.8397, demonstrating that ensembling diverse models can lead to reliable classification results, particularly in scenarios involving multi-class data.

Our CNN-based model with data augmentation (CNN-AUG) stood out significantly in terms of performance. It achieved an outstanding accuracy of 0.9856, with precision, recall, and F1-scores all reaching 0.9877. These near-perfect scores reflect the model’s ability to correctly identify and classify instances across all classes with minimal errors. Additionally, the CNN-AUG model achieved an AUC of 1.00, indicating flawless discriminative capability between classes. Similarly, the AP score of 1.00 emphasized the model's exceptional precision across varying thresholds, underscoring its effectiveness in capturing distinguishing features and nuances within the dataset. The implementation of data augmentation techniques played a crucial role in enhancing the CNN’s capacity to generalize well on unseen data, effectively addressing class imbalances and overfitting issues commonly encountered in deep learning models.
In conclusion, our comparative analysis highlights the superiority of the CNN-AUG model over traditional machine learning approaches. While traditional models such as Random Forest and the Voting Classifier demonstrated commendable performance, particularly in achieving high precision and robust AUC scores, the CNN-AUG model outperformed them across all evaluation metrics. The integration of convolutional neural network architecture with data augmentation techniques allowed the CNN-AUG model to learn complex patterns and relationships within the data effectively, resulting in near-perfect accuracy and reliability in multi-class classification.
\begin{figure}
    \centering
    \includegraphics[width=0.8\textwidth]{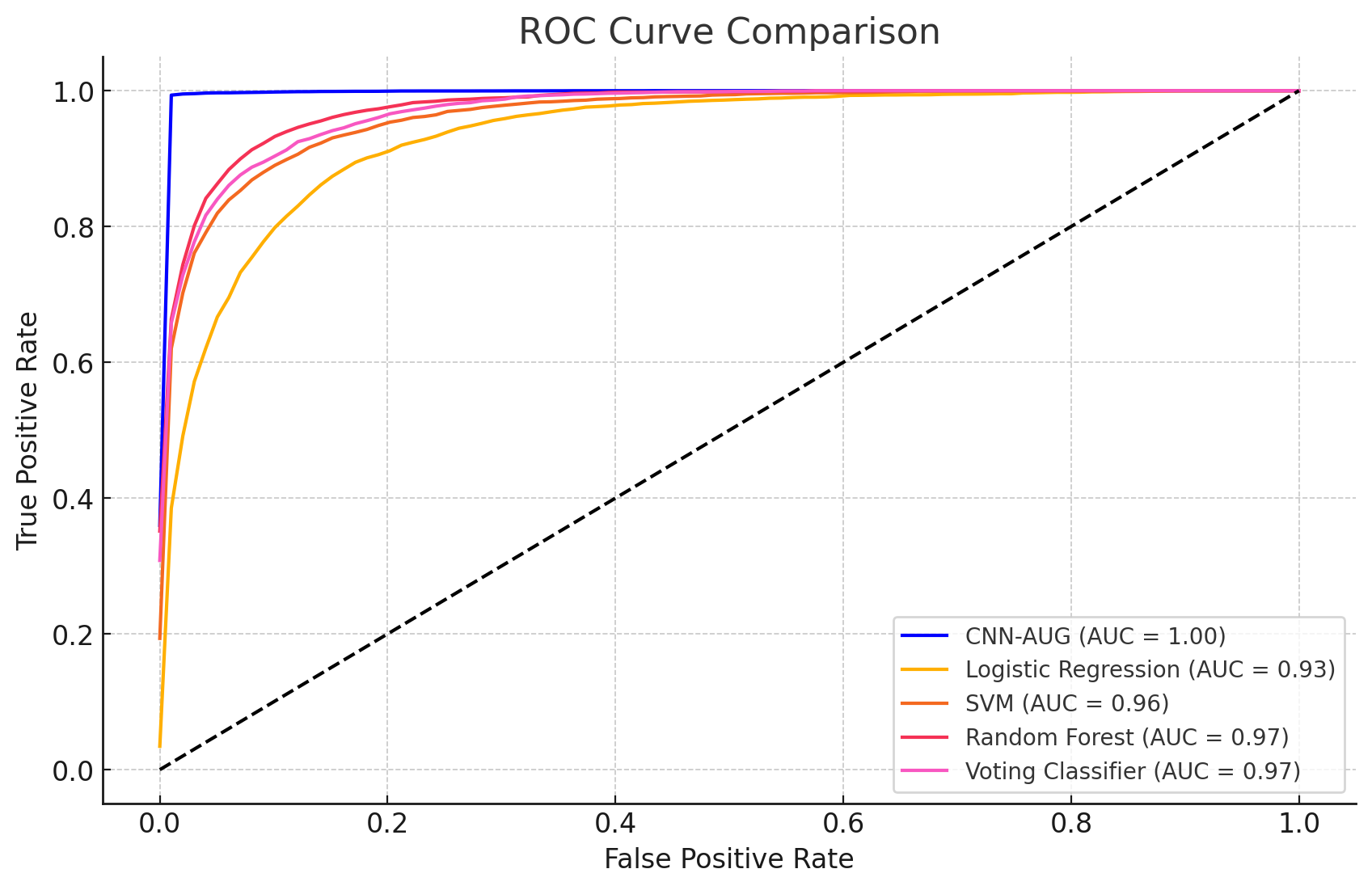}
    \caption{ ROC Curve Comparison.
The ROC curve compares the performance of various models, including CNN-AUG, SVC, and Random Forest, on the test dataset. The curves show the trade-off between the True Positive Rate (TPR) and the False Positive Rate (FPR) for each model. The Area Under the Curve (AUC) values are provided in the legend, indicating that the CNN-AUG model has the highest AUC, demonstrating superior classification performance in distinguishing between the classes.}
    \label{fig:8}
\end{figure}

\begin{figure}
    \centering
    \includegraphics[width=0.8\textwidth]{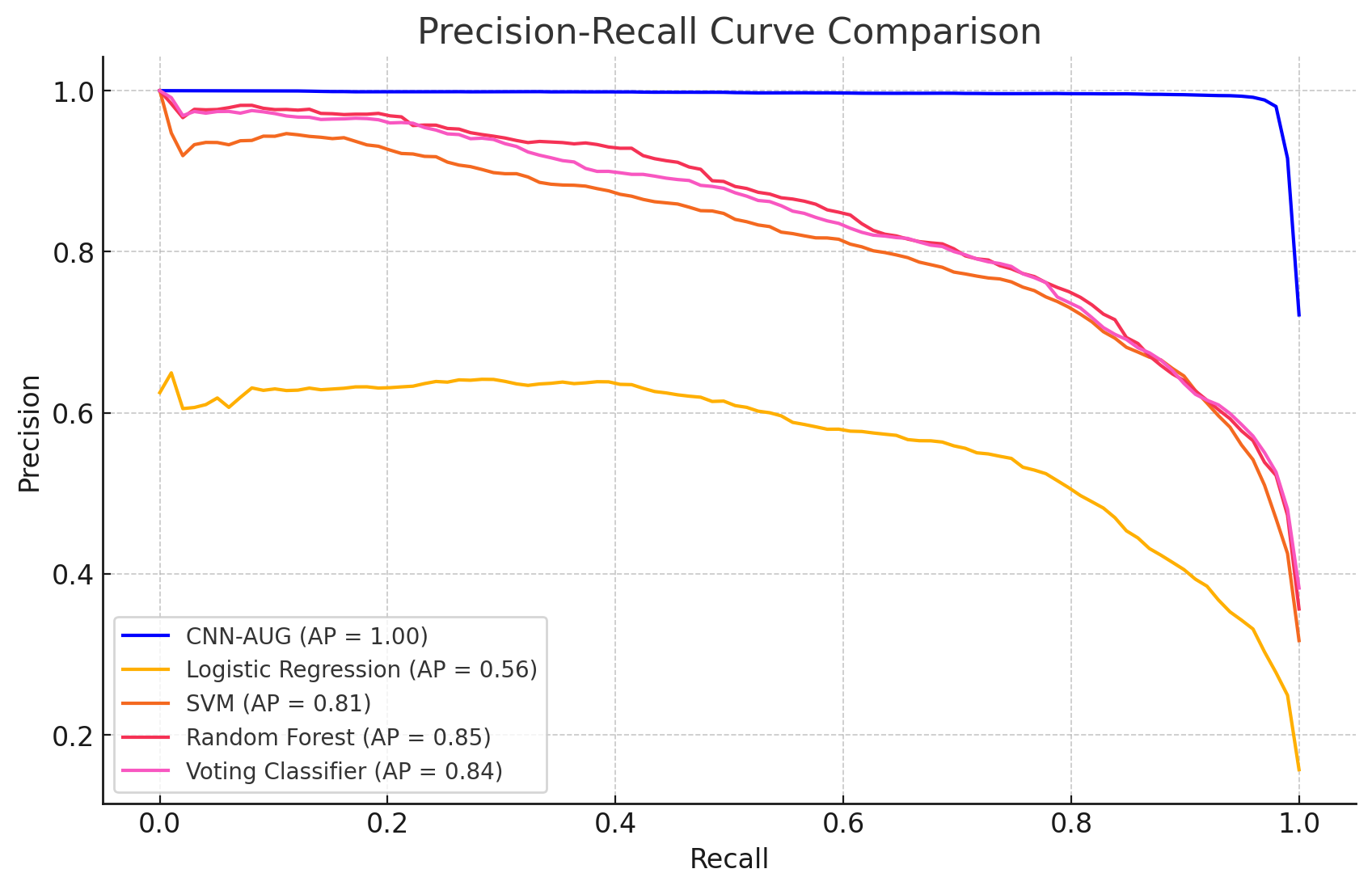}
    \caption{Precision-Recall Curve Comparison.
The Precision-Recall curve evaluates the precision against recall for different models on the test dataset. The graph highlights the trade-offs between achieving high precision and high recall for each model, with Average Precision (AP) values provided in the legend. The CNN-AUG model achieves a higher AP, indicating its better ability to maintain high precision while achieving higher recall compared to other models like SVC and Random Forest.}
    \label{fig:9}
\end{figure}

\begin{table}[t]
    \centering
    \scriptsize
    \caption{Model Performance Comparison}
    \setlength{\tabcolsep}{1pt} 
    \begin{tabular*}{0.8\textwidth}{@{\extracolsep{\fill}}ccccccc@{}}
        \toprule
        Model & Acc. & Prec. & Recall & F1 & AUC & AP \\
        \midrule
        Logistic Regression & 0.7176 & 0.5172 & 0.5147 & 0.5137 & 0.9345 & 0.5653 \\
        SVM &  0.7770 & 0.7627 & 0.7244 & 0.7283 & 0.9638 & 0.8141 \\
        Random Forest & 0.8125 & 0.8482 & 0.7090 & 0.7318 & 0.9759 & 0.8476 \\
        Voting Classifier & 0.7951 & 0.8050 & 0.7322 & 0.7519 & 0.9709 & 0.8397 \\
        \textbf{CNN-AUG} & \textbf{0.9856} & \textbf{0.9877} & \textbf{0.9878} & \textbf{0.9877} & \textbf{1.0000} & \textbf{1.0000} \\
        \bottomrule
    \end{tabular*}
    \label{tab:model_performance}
\end{table}

\subsection{Ablation Analysis to Assess Component Contributions}
We developed a series of Convolutional Neural Network (CNN) models to investigate the contribution of different layers to classification performance. The baseline model (CNN-AUG) consists of three convolutional layers. The first convolutional layer applies 16 filters of size 3×33×3 with the ReLU activation function and same padding. The second and third convolutional layers utilize 64 and 128 filters, respectively, with similar configurations. The output from the third convolutional layer is flattened and fed into two fully-connected layers with 512 and 128 units, both using ReLU activations. The final output layer has 8 units with a softmax activation function for multi-class classification.
To understand the effect of different architectural components, we created two ablation models. The first ablation model, named No Conv3, excludes the third convolutional layer, keeping the rest of the architecture unchanged. The second ablation model, named No Dense1, excludes the first fully-connected (Dense) layer, retaining only the second fully-connected layer. Each model was trained using the Adam optimizer with a categorical cross-entropy loss function.
Results
We evaluated the models on the test dataset and compared their performance using accuracy, precision, recall, and F1 score. The results, as shown in Table \ref{tab:ablation_performance}, indicate that the baseline model (CNN-AUG) achieved the highest performance across all metrics, with a test accuracy of 0.9856, precision of 0.9855, recall of 0.9856, and an F1 score of 0.9855. This confirms the model's effectiveness in accurately classifying the eight classes in the dataset.

The No Conv3 model, which excludes the third convolutional layer, showed a slight drop in performance. It achieved a test accuracy of 0.9796, precision of 0.9795, recall of 0.9795, and an F1 score of 0.9794. These results suggest that the third convolutional layer contributes to deeper feature extraction, enhancing the model’s overall performance.

Similarly, the No Dense1 model, which removes the first fully-connected (dense) layer, also exhibited a decline in all metrics. It achieved a test accuracy of 0.9784, precision of 0.9785, recall of 0.9785, and an F1 score of 0.9784. This indicates that having multiple dense layers is crucial for effective classification, as it allows the model to better capture and learn the complex relationships between features.

These results collectively highlight the importance of both the third convolutional layer and the first fully-connected layer in achieving optimal classification performance.
\begin{table}[t]
    \centering
    \caption{Ablation Study Results}
    \begin{tabular}{lcccc}
        \toprule
        Model & Acc & Prec & Rec & F1 \\
        \midrule
        \textbf{CNN-AUG} & 0.9856 & 0.9855 & 0.9856 & 0.9855 \\
        No Conv3 & 0.9796 & 0.9795 & 0.9795 & 0.9794 \\
        No Dense1 & 0.9784 & 0.9785 & 0.9785 & 0.9784 \\
        \bottomrule
    \end{tabular}
    \label{tab:ablation_performance}
\end{table}

\subsection{Interpretable Exploration through Regional Occlusion Analysis}

\begin{figure}[h]
    \centering
    \includegraphics[width=0.8\textwidth]{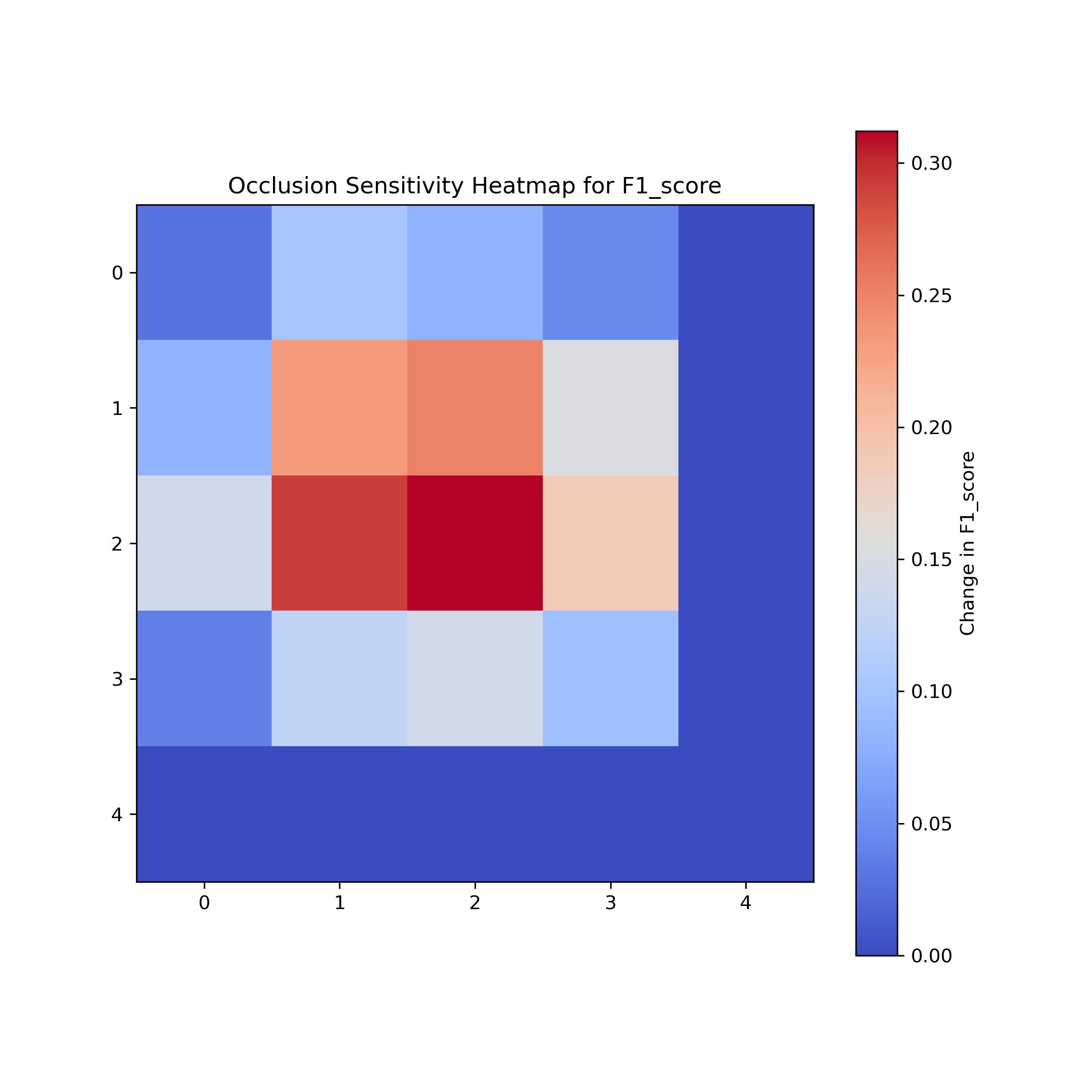}
    \caption{Occlusion Sensitivity Heatmap illustrating the change in F1 score due to masking different regions of the wafer image. The color bar represents the magnitude of change in the F1 score, indicating the importance of different regions for model prediction.}
    \label{fig:11}
\end{figure}

In this experiment, we performed an occlusion sensitivity analysis to explain the performance of our deep learning model in a wafer defect classification task. The main goal is to identify spatial regions in the wafer image that are critical for the model's decision-making process. By systematically occluding image regions using a sliding window of 10x10 pixels with a spacing of 5 pixels, we evaluated the impact of occlusion on the model's classification performance, with a particular focus on the F1 score as a balanced indicator of precision and recall. The analysis generates a heatmap as shown in Fig ~\ref{fig:11}, where the center regions of the wafer images often contain prominent defective features, and masking these regions leads to a significant decrease in the F1 score. This finding highlights the dependence of the model on the center region for accurate classification.
These results have practical implications for both model design and wafer detection strategies. The analysis highlights the importance of central regions in feature extraction, providing insight into how deep learning models prioritize spatial information. Understanding these critical regions allows for refinement of model architectures, such as designing mechanisms that more effectively focus on relevant image regions. In addition, the improved interpretability obtained through occlusion sensitivity analysis enhances the trust in deep learning models and provides a way to optimize the performance and robustness of the models in real-world applications.

\section{Conclusions}
In this paper, we proposed a novel approach combining self-encoder-based data enhancement with convolutional neural networks (CNNs) to improve the classification accuracy of wafer defect maps (WDMs). By introducing noise in the latent space of the self-encoder, we augmented the diversity of training data and mitigated class imbalance, enabling the CNN to extract hierarchical features for precise defect classification. Experiments conducted on the WM-811K dataset demonstrated the superiority of the proposed method, achieving a classification accuracy of 98.56\%, significantly outperforming traditional machine learning methods. These results validate the effectiveness of our approach in addressing the challenges of noisy data, unbalanced defect classes, and complex failure patterns.

Looking forward, two key areas of improvement need to be addressed. First, the current method requires resizing wafer maps to a uniform shape, which may distort certain defect patterns and impact classification accuracy for wafer maps with extreme aspect ratios. Future work will explore techniques to handle arbitrarily-shaped wafer maps without resizing. Second, while the overall classification performance improved, certain minority defect categories, such as Scratch and Near-Full, still suffered from lower accuracy. To address this, future efforts will focus on adaptive and class-specific data augmentation strategies and cost-sensitive learning methods to enhance classification performance for these critical but underrepresented categories.

\vspace{\baselineskip}
\textbf{Acknowledgment}: This work was supported by  National Natural Science Foundation of China, “Research on Multiplicity Object Classification for Heterogeneous Marker Space”(Grant No. 62306131).

\textbf{Conflict of Interest}: The authors declare that there is no conflict of interests regarding the publication of this article.

\textbf{Data availability}: Data will be made available on reasonable request.

\bibliographystyle{sn-mathphys} 


\begin{thebibliography}{37}
\ifx \bisbn   \undefined \def \bisbn  #1{ISBN #1}\fi
\ifx \binits  \undefined \def \binits#1{#1}\fi
\ifx \bauthor  \undefined \def \bauthor#1{#1}\fi
\ifx \batitle  \undefined \def \batitle#1{#1}\fi
\ifx \bjtitle  \undefined \def \bjtitle#1{#1}\fi
\ifx \bvolume  \undefined \def \bvolume#1{\textbf{#1}}\fi
\ifx \byear  \undefined \def \byear#1{#1}\fi
\ifx \bissue  \undefined \def \bissue#1{#1}\fi
\ifx \bfpage  \undefined \def \bfpage#1{#1}\fi
\ifx \blpage  \undefined \def \blpage #1{#1}\fi
\ifx \burl  \undefined \def \burl#1{\textsf{#1}}\fi
\ifx \doiurl  \undefined \def \doiurl#1{\url{https://doi.org/#1}}\fi
\ifx \betal  \undefined \def \betal{\textit{et al.}}\fi
\ifx \binstitute  \undefined \def \binstitute#1{#1}\fi
\ifx \binstitutionaled  \undefined \def \binstitutionaled#1{#1}\fi
\ifx \bctitle  \undefined \def \bctitle#1{#1}\fi
\ifx \beditor  \undefined \def \beditor#1{#1}\fi
\ifx \bpublisher  \undefined \def \bpublisher#1{#1}\fi
\ifx \bbtitle  \undefined \def \bbtitle#1{#1}\fi
\ifx \bedition  \undefined \def \bedition#1{#1}\fi
\ifx \bseriesno  \undefined \def \bseriesno#1{#1}\fi
\ifx \blocation  \undefined \def \blocation#1{#1}\fi
\ifx \bsertitle  \undefined \def \bsertitle#1{#1}\fi
\ifx \bsnm \undefined \def \bsnm#1{#1}\fi
\ifx \bsuffix \undefined \def \bsuffix#1{#1}\fi
\ifx \bparticle \undefined \def \bparticle#1{#1}\fi
\ifx \barticle \undefined \def \barticle#1{#1}\fi
\bibcommenthead
\ifx \bconfdate \undefined \def \bconfdate #1{#1}\fi
\ifx \botherref \undefined \def \botherref #1{#1}\fi
\ifx \url \undefined \def \url#1{\textsf{#1}}\fi
\ifx \bchapter \undefined \def \bchapter#1{#1}\fi
\ifx \bbook \undefined \def \bbook#1{#1}\fi
\ifx \bcomment \undefined \def \bcomment#1{#1}\fi
\ifx \oauthor \undefined \def \oauthor#1{#1}\fi
\ifx \citeauthoryear \undefined \def \citeauthoryear#1{#1}\fi
\ifx \endbibitem  \undefined \def \endbibitem {}\fi
\ifx \bconflocation  \undefined \def \bconflocation#1{#1}\fi
\ifx \arxivurl  \undefined \def \arxivurl#1{\textsf{#1}}\fi
\csname PreBibitemsHook\endcsname

\bibitem[\protect\citeauthoryear{Wei et~al.}{2023}]{wei2023wafer}
\begin{bchapter}
\bauthor{\bsnm{Wei}, \binits{Q.}},
\bauthor{\bsnm{Zhao}, \binits{W.}},
\bauthor{\bsnm{Zheng}, \binits{X.}},
\bauthor{\bsnm{Zeng}, \binits{Z.}}:
\bctitle{Wafer map defect patterns semi-supervised classification using latent vector representation}.
In: \bbtitle{2023 IEEE International Conference on Cybernetics and Intelligent Systems (CIS) and IEEE Conference on Robotics, Automation and Mechatronics (RAM)},
pp. \bfpage{192}--\blpage{197}
(\byear{2023}).
\bcomment{IEEE}
\end{bchapter}
\endbibitem

\bibitem[\protect\citeauthoryear{De~Ridder et~al.}{2023}]{de2023semi}
\begin{bchapter}
\bauthor{\bsnm{De~Ridder}, \binits{V.}},
\bauthor{\bsnm{Dey}, \binits{B.}},
\bauthor{\bsnm{Dehaerne}, \binits{E.}},
\bauthor{\bsnm{Halder}, \binits{S.}},
\bauthor{\bsnm{De~Gendt}, \binits{S.}},
\bauthor{\bsnm{Van~Waeyenberge}, \binits{B.}}:
\bctitle{Semi-centernet: a machine learning facilitated approach for semiconductor defect inspection}.
In: \bbtitle{38th European Mask and Lithography Conference (EMLC 2023)},
vol. \bseriesno{12802},
pp. \bfpage{220}--\blpage{228}
(\byear{2023}).
\bcomment{SPIE}
\end{bchapter}
\endbibitem

\bibitem[\protect\citeauthoryear{Yin et~al.}{2019}]{yin2019stone}
\begin{barticle}
\bauthor{\bsnm{Yin}, \binits{H.}},
\bauthor{\bsnm{Shi}, \binits{X.}},
\bauthor{\bsnm{He}, \binits{C.}},
\bauthor{\bsnm{Martinez-Canales}, \binits{M.}},
\bauthor{\bsnm{Li}, \binits{J.}},
\bauthor{\bsnm{Pickard}, \binits{C.J.}},
\bauthor{\bsnm{Tang}, \binits{C.}},
\bauthor{\bsnm{Ouyang}, \binits{T.}},
\bauthor{\bsnm{Zhang}, \binits{C.}},
\bauthor{\bsnm{Zhong}, \binits{J.}}:
\batitle{Stone-wales graphene: A two-dimensional carbon semimetal with magic stability}.
\bjtitle{Physical Review B}
\bvolume{99}(\bissue{4}),
\bfpage{041405}
(\byear{2019})
\end{barticle}
\endbibitem

\bibitem[\protect\citeauthoryear{Mills and Le~Hunte}{1997}]{mills1997overview}
\begin{barticle}
\bauthor{\bsnm{Mills}, \binits{A.}},
\bauthor{\bsnm{Le~Hunte}, \binits{S.}}:
\batitle{An overview of semiconductor photocatalysis}.
\bjtitle{Journal of photochemistry and photobiology A: Chemistry}
\bvolume{108}(\bissue{1}),
\bfpage{1}--\blpage{35}
(\byear{1997})
\end{barticle}
\endbibitem

\bibitem[\protect\citeauthoryear{Yoon and Kang}{2022}]{yoon2022semi}
\begin{barticle}
\bauthor{\bsnm{Yoon}, \binits{S.}},
\bauthor{\bsnm{Kang}, \binits{S.}}:
\batitle{Semi-automatic wafer map pattern classification with convolutional neural networks}.
\bjtitle{Computers \& Industrial Engineering}
\bvolume{166},
\bfpage{107977}
(\byear{2022})
\end{barticle}
\endbibitem

\bibitem[\protect\citeauthoryear{Ishida et~al.}{2019}]{ishida2019deep}
\begin{bchapter}
\bauthor{\bsnm{Ishida}, \binits{T.}},
\bauthor{\bsnm{Nitta}, \binits{I.}},
\bauthor{\bsnm{Fukuda}, \binits{D.}},
\bauthor{\bsnm{Kanazawa}, \binits{Y.}}:
\bctitle{Deep learning-based wafer-map failure pattern recognition framework}.
In: \bbtitle{20th International Symposium on Quality Electronic Design (ISQED)},
pp. \bfpage{291}--\blpage{297}
(\byear{2019}).
\bcomment{IEEE}
\end{bchapter}
\endbibitem

\bibitem[\protect\citeauthoryear{Ferris-Prabhu}{1989}]{ferris1989defects}
\begin{bchapter}
\bauthor{\bsnm{Ferris-Prabhu}, \binits{A.V.}}:
\bctitle{Defects, faults and semiconductor device yield}.
In: \beditor{\bsnm{Koren}, \binits{I.}} (ed.)
\bbtitle{Defect and Fault Tolerance in VLSI Systems: Volume 1},
pp. \bfpage{33}--\blpage{46}.
\bpublisher{Springer},
\blocation{Boston, MA}
(\byear{1989})
\end{bchapter}
\endbibitem

\bibitem[\protect\citeauthoryear{Poehls et~al.}{2021}]{poehls2021review}
\begin{barticle}
\bauthor{\bsnm{Poehls}, \binits{L.B.}},
\bauthor{\bsnm{Fieback}, \binits{M.}},
\bauthor{\bsnm{Hoffmann-Eifert}, \binits{S.}},
\bauthor{\bsnm{Copetti}, \binits{T.}},
\bauthor{\bsnm{Brum}, \binits{E.}},
\bauthor{\bsnm{Menzel}, \binits{S.}},
\bauthor{\bsnm{Hamdioui}, \binits{S.}},
\bauthor{\bsnm{Gemmeke}, \binits{T.}}:
\batitle{Review of manufacturing process defects and their effects on memristive devices}.
\bjtitle{Journal of electronic testing}
\bvolume{37},
\bfpage{427}--\blpage{437}
(\byear{2021})
\end{barticle}
\endbibitem

\bibitem[\protect\citeauthoryear{Kim et~al.}{2021}]{kim2021adversarial}
\begin{barticle}
\bauthor{\bsnm{Kim}, \binits{J.}},
\bauthor{\bsnm{Nam}, \binits{Y.}},
\bauthor{\bsnm{Kang}, \binits{M.-C.}},
\bauthor{\bsnm{Kim}, \binits{K.}},
\bauthor{\bsnm{Hong}, \binits{J.}},
\bauthor{\bsnm{Lee}, \binits{S.}},
\bauthor{\bsnm{Kim}, \binits{D.-N.}}:
\batitle{Adversarial defect detection in semiconductor manufacturing process}.
\bjtitle{IEEE Transactions on Semiconductor Manufacturing}
\bvolume{34}(\bissue{3}),
\bfpage{365}--\blpage{371}
(\byear{2021})
\end{barticle}
\endbibitem

\bibitem[\protect\citeauthoryear{Maksim et~al.}{2019}]{maksim2019classification}
\begin{bchapter}
\bauthor{\bsnm{Maksim}, \binits{K.}},
\bauthor{\bsnm{Kirill}, \binits{B.}},
\bauthor{\bsnm{Eduard}, \binits{Z.}},
\bauthor{\bsnm{Nikita}, \binits{G.}},
\bauthor{\bsnm{Aleksandr}, \binits{B.}},
\bauthor{\bsnm{Arina}, \binits{L.}},
\bauthor{\bsnm{Vladislav}, \binits{S.}},
\bauthor{\bsnm{Daniil}, \binits{M.}},
\bauthor{\bsnm{Nikolay}, \binits{K.}}:
\bctitle{Classification of wafer maps defect based on deep learning methods with small amount of data}.
In: \bbtitle{2019 International Conference on Engineering and Telecommunication (EnT)},
pp. \bfpage{1}--\blpage{5}
(\byear{2019}).
\bcomment{IEEE}
\end{bchapter}
\endbibitem

\bibitem[\protect\citeauthoryear{Shim and Kang}{2023}]{shim2023learning}
\begin{barticle}
\bauthor{\bsnm{Shim}, \binits{J.}},
\bauthor{\bsnm{Kang}, \binits{S.}}:
\batitle{Learning from single-defect wafer maps to classify mixed-defect wafer maps}.
\bjtitle{Expert Systems with Applications}
\bvolume{233},
\bfpage{120923}
(\byear{2023})
\end{barticle}
\endbibitem

\bibitem[\protect\citeauthoryear{Hansen and Thyregod}{1998}]{hansen1998use}
\begin{barticle}
\bauthor{\bsnm{Hansen}, \binits{C.}},
\bauthor{\bsnm{Thyregod}, \binits{P.}}:
\batitle{Use of wafer maps in integrated circuit manufacturing}.
\bjtitle{Microelectronics Reliability}
\bvolume{38}(\bissue{6-8}),
\bfpage{1155}--\blpage{1164}
(\byear{1998})
\end{barticle}
\endbibitem

\bibitem[\protect\citeauthoryear{Kang et~al.}{2015}]{kang2015using}
\begin{barticle}
\bauthor{\bsnm{Kang}, \binits{S.}},
\bauthor{\bsnm{Cho}, \binits{S.}},
\bauthor{\bsnm{An}, \binits{D.}},
\bauthor{\bsnm{Rim}, \binits{J.}}:
\batitle{Using wafer map features to better predict die-level failures in final test}.
\bjtitle{IEEE Transactions on Semiconductor Manufacturing}
\bvolume{28}(\bissue{3}),
\bfpage{431}--\blpage{437}
(\byear{2015})
\end{barticle}
\endbibitem

\bibitem[\protect\citeauthoryear{Taha et~al.}{2017}]{taha2017clustering}
\begin{barticle}
\bauthor{\bsnm{Taha}, \binits{K.}},
\bauthor{\bsnm{Salah}, \binits{K.}},
\bauthor{\bsnm{Yoo}, \binits{P.D.}}:
\batitle{Clustering the dominant defective patterns in semiconductor wafer maps}.
\bjtitle{IEEE Transactions on Semiconductor Manufacturing}
\bvolume{31}(\bissue{1}),
\bfpage{156}--\blpage{165}
(\byear{2017})
\end{barticle}
\endbibitem

\bibitem[\protect\citeauthoryear{Alawieh et~al.}{2017}]{alawieh2017identifying}
\begin{barticle}
\bauthor{\bsnm{Alawieh}, \binits{M.B.}},
\bauthor{\bsnm{Wang}, \binits{F.}},
\bauthor{\bsnm{Li}, \binits{X.}}:
\batitle{Identifying wafer-level systematic failure patterns via unsupervised learning}.
\bjtitle{IEEE transactions on computer-aided design of integrated circuits and systems}
\bvolume{37}(\bissue{4}),
\bfpage{832}--\blpage{844}
(\byear{2017})
\end{barticle}
\endbibitem

\bibitem[\protect\citeauthoryear{Geng et~al.}{2023}]{geng2023mixed}
\begin{bchapter}
\bauthor{\bsnm{Geng}, \binits{H.}},
\bauthor{\bsnm{Sun}, \binits{Q.}},
\bauthor{\bsnm{Chen}, \binits{T.}},
\bauthor{\bsnm{Xu}, \binits{Q.}},
\bauthor{\bsnm{Ho}, \binits{T.-Y.}},
\bauthor{\bsnm{Yu}, \binits{B.}}:
\bctitle{Mixed-type wafer failure pattern recognition}.
In: \bbtitle{Proceedings of the 28th Asia and South Pacific Design Automation Conference},
pp. \bfpage{727}--\blpage{732}
(\byear{2023})
\end{bchapter}
\endbibitem

\bibitem[\protect\citeauthoryear{Nakazawa and Kulkarni}{2018}]{nakazawa2018wafer}
\begin{barticle}
\bauthor{\bsnm{Nakazawa}, \binits{T.}},
\bauthor{\bsnm{Kulkarni}, \binits{D.V.}}:
\batitle{Wafer map defect pattern classification and image retrieval using convolutional neural network}.
\bjtitle{IEEE Transactions on Semiconductor Manufacturing}
\bvolume{31}(\bissue{2}),
\bfpage{309}--\blpage{314}
(\byear{2018})
\end{barticle}
\endbibitem

\bibitem[\protect\citeauthoryear{Piao and Jin}{2023}]{piao2023cnn}
\begin{barticle}
\bauthor{\bsnm{Piao}, \binits{M.}},
\bauthor{\bsnm{Jin}, \binits{C.H.}}:
\batitle{Cnn and ensemble learning based wafer map failure pattern recognition based on local property based features}.
\bjtitle{Journal of Intelligent Manufacturing}
\bvolume{34}(\bissue{8}),
\bfpage{3599}--\blpage{3621}
(\byear{2023})
\end{barticle}
\endbibitem

\bibitem[\protect\citeauthoryear{Sumikawa et~al.}{2013}]{sumikawa2013pattern}
\begin{bchapter}
\bauthor{\bsnm{Sumikawa}, \binits{N.}},
\bauthor{\bsnm{Wang}, \binits{L.-C.}},
\bauthor{\bsnm{Abadir}, \binits{M.S.}}:
\bctitle{A pattern mining framework for inter-wafer abnormality analysis}.
In: \bbtitle{2013 IEEE International Test Conference (ITC)},
pp. \bfpage{1}--\blpage{10}
(\byear{2013}).
\bcomment{IEEE}
\end{bchapter}
\endbibitem

\bibitem[\protect\citeauthoryear{Hearst et~al.}{1998}]{hearst1998support}
\begin{barticle}
\bauthor{\bsnm{Hearst}, \binits{M.A.}},
\bauthor{\bsnm{Dumais}, \binits{S.T.}},
\bauthor{\bsnm{Osuna}, \binits{E.}},
\bauthor{\bsnm{Platt}, \binits{J.}},
\bauthor{\bsnm{Scholkopf}, \binits{B.}}:
\batitle{Support vector machines}.
\bjtitle{IEEE Intelligent Systems and their applications}
\bvolume{13}(\bissue{4}),
\bfpage{18}--\blpage{28}
(\byear{1998})
\end{barticle}
\endbibitem

\bibitem[\protect\citeauthoryear{Peterson}{2009}]{peterson2009k}
\begin{barticle}
\bauthor{\bsnm{Peterson}, \binits{L.E.}}:
\batitle{K-nearest neighbor}.
\bjtitle{Scholarpedia}
\bvolume{4}(\bissue{2}),
\bfpage{1883}
(\byear{2009})
\end{barticle}
\endbibitem

\bibitem[\protect\citeauthoryear{De~Ville}{2013}]{de2013decision}
\begin{barticle}
\bauthor{\bsnm{De~Ville}, \binits{B.}}:
\batitle{Decision trees}.
\bjtitle{Wiley Interdisciplinary Reviews: Computational Statistics}
\bvolume{5}(\bissue{6}),
\bfpage{448}--\blpage{455}
(\byear{2013})
\end{barticle}
\endbibitem

\bibitem[\protect\citeauthoryear{Wu et~al.}{2014}]{wu2014wafer}
\begin{barticle}
\bauthor{\bsnm{Wu}, \binits{M.-J.}},
\bauthor{\bsnm{Jang}, \binits{J.-S.R.}},
\bauthor{\bsnm{Chen}, \binits{J.-L.}}:
\batitle{Wafer map failure pattern recognition and similarity ranking for large-scale data sets}.
\bjtitle{IEEE Transactions on Semiconductor Manufacturing}
\bvolume{28}(\bissue{1}),
\bfpage{1}--\blpage{12}
(\byear{2014})
\end{barticle}
\endbibitem

\bibitem[\protect\citeauthoryear{Yin et~al.}{2024a}]{yin2024evaluation}
\begin{botherref}
\oauthor{\bsnm{Yin}, \binits{H.}},
\oauthor{\bsnm{Gu}, \binits{Z.}},
\oauthor{\bsnm{Wang}, \binits{F.}},
\oauthor{\bsnm{Abuduhaibaier}, \binits{Y.}},
\oauthor{\bsnm{Zhu}, \binits{Y.}},
\oauthor{\bsnm{Tu}, \binits{X.}},
\oauthor{\bsnm{Hua}, \binits{X.-S.}},
\oauthor{\bsnm{Luo}, \binits{X.}},
\oauthor{\bsnm{Sun}, \binits{Y.}}:
An evaluation of large language models in bioinformatics research.
arXiv preprint arXiv:2402.13714
(2024)
\end{botherref}
\endbibitem

\bibitem[\protect\citeauthoryear{Yin et~al.}{2024b}]{yin2024ipev}
\begin{barticle}
\bauthor{\bsnm{Yin}, \binits{H.}},
\bauthor{\bsnm{Wu}, \binits{S.}},
\bauthor{\bsnm{Tan}, \binits{J.}},
\bauthor{\bsnm{Guo}, \binits{Q.}},
\bauthor{\bsnm{Li}, \binits{M.}},
\bauthor{\bsnm{Guo}, \binits{J.}},
\bauthor{\bsnm{Wang}, \binits{Y.}},
\bauthor{\bsnm{Jiang}, \binits{X.}},
\bauthor{\bsnm{Zhu}, \binits{H.}}:
\batitle{Ipev: identification of prokaryotic and eukaryotic virus-derived sequences in virome using deep learning}.
\bjtitle{GigaScience}
\bvolume{13},
\bfpage{018}
(\byear{2024})
\end{barticle}
\endbibitem

\bibitem[\protect\citeauthoryear{Wang and Chen}{2020}]{wang2020defect}
\begin{barticle}
\bauthor{\bsnm{Wang}, \binits{R.}},
\bauthor{\bsnm{Chen}, \binits{N.}}:
\batitle{Defect pattern recognition on wafers using convolutional neural networks}.
\bjtitle{Quality and Reliability Engineering International}
\bvolume{36}(\bissue{4}),
\bfpage{1245}--\blpage{1257}
(\byear{2020})
\end{barticle}
\endbibitem

\bibitem[\protect\citeauthoryear{Phua and Theng}{2020}]{phua2020semiconductor}
\begin{bchapter}
\bauthor{\bsnm{Phua}, \binits{C.}},
\bauthor{\bsnm{Theng}, \binits{L.B.}}:
\bctitle{Semiconductor wafer surface: Automatic defect classification with deep cnn}.
In: \bbtitle{2020 IEEE Region 10 Conference (TENCON)},
pp. \bfpage{714}--\blpage{719}
(\byear{2020}).
\bcomment{IEEE}
\end{bchapter}
\endbibitem

\bibitem[\protect\citeauthoryear{Yu et~al.}{2019}]{yu2019stacked}
\begin{barticle}
\bauthor{\bsnm{Yu}, \binits{J.}},
\bauthor{\bsnm{Zheng}, \binits{X.}},
\bauthor{\bsnm{Liu}, \binits{J.}}:
\batitle{Stacked convolutional sparse denoising auto-encoder for identification of defect patterns in semiconductor wafer map}.
\bjtitle{Computers in Industry}
\bvolume{109},
\bfpage{121}--\blpage{133}
(\byear{2019})
\end{barticle}
\endbibitem

\bibitem[\protect\citeauthoryear{Liou et~al.}{2014}]{liou2014autoencoder}
\begin{barticle}
\bauthor{\bsnm{Liou}, \binits{C.-Y.}},
\bauthor{\bsnm{Cheng}, \binits{W.-C.}},
\bauthor{\bsnm{Liou}, \binits{J.-W.}},
\bauthor{\bsnm{Liou}, \binits{D.-R.}}:
\batitle{Autoencoder for words}.
\bjtitle{Neurocomputing}
\bvolume{139},
\bfpage{84}--\blpage{96}
(\byear{2014})
\end{barticle}
\endbibitem

\bibitem[\protect\citeauthoryear{Zhai et~al.}{2018}]{zhai2018autoencoder}
\begin{bchapter}
\bauthor{\bsnm{Zhai}, \binits{J.}},
\bauthor{\bsnm{Zhang}, \binits{S.}},
\bauthor{\bsnm{Chen}, \binits{J.}},
\bauthor{\bsnm{He}, \binits{Q.}}:
\bctitle{Autoencoder and its various variants}.
In: \bbtitle{2018 IEEE International Conference on Systems, Man, and Cybernetics (SMC)},
pp. \bfpage{415}--\blpage{419}
(\byear{2018}).
\bcomment{IEEE}
\end{bchapter}
\endbibitem

\bibitem[\protect\citeauthoryear{Cheon et~al.}{2019}]{cheon2019convolutional}
\begin{barticle}
\bauthor{\bsnm{Cheon}, \binits{S.}},
\bauthor{\bsnm{Lee}, \binits{H.}},
\bauthor{\bsnm{Kim}, \binits{C.O.}},
\bauthor{\bsnm{Lee}, \binits{S.H.}}:
\batitle{Convolutional neural network for wafer surface defect classification and the detection of unknown defect class}.
\bjtitle{IEEE Transactions on Semiconductor Manufacturing}
\bvolume{32}(\bissue{2}),
\bfpage{163}--\blpage{170}
(\byear{2019})
\end{barticle}
\endbibitem

\bibitem[\protect\citeauthoryear{Kim and Kim}{2023}]{kim2023mixed}
\begin{botherref}
\oauthor{\bsnm{Kim}, \binits{S.}},
\oauthor{\bsnm{Kim}, \binits{H.}}:
Mixed-type defect pattern recognition in noisy labeled wafer bin maps.
Quality Engineering,
1--15
(2023)
\end{botherref}
\endbibitem

\bibitem[\protect\citeauthoryear{Bhatnagar et~al.}{2022}]{bhatnagar2022semiconductor}
\begin{bchapter}
\bauthor{\bsnm{Bhatnagar}, \binits{P.}},
\bauthor{\bsnm{Arora}, \binits{T.}},
\bauthor{\bsnm{Chaujar}, \binits{R.}}:
\bctitle{Semiconductor wafer map defect classification using transfer learning}.
In: \bbtitle{2022 IEEE Delhi Section Conference (DELCON)},
pp. \bfpage{1}--\blpage{4}
(\byear{2022}).
\bcomment{IEEE}
\end{bchapter}
\endbibitem

\bibitem[\protect\citeauthoryear{Kane}{1996}]{kane1996precision}
\begin{barticle}
\bauthor{\bsnm{Kane}, \binits{M.}}:
\batitle{The precision of measurements}.
\bjtitle{Applied measurement in education}
\bvolume{9}(\bissue{4}),
\bfpage{355}--\blpage{379}
(\byear{1996})
\end{barticle}
\endbibitem

\bibitem[\protect\citeauthoryear{Van~Rossum and Drake~Jr}{1995}]{van1995python}
\begin{bbook}
\bauthor{\bsnm{Van~Rossum}, \binits{G.}},
\bauthor{\bsnm{Drake~Jr}, \binits{F.L.}}:
\bbtitle{Python Tutorial}
vol. \bseriesno{620},
\bedition{1st} edn.
\bpublisher{Centrum voor Wiskunde en Informatica},
\blocation{Amsterdam, The Netherlands}
(\byear{1995})
\end{bbook}
\endbibitem

\bibitem[\protect\citeauthoryear{Pang et~al.}{2020}]{pang2020deep}
\begin{barticle}
\bauthor{\bsnm{Pang}, \binits{B.}},
\bauthor{\bsnm{Nijkamp}, \binits{E.}},
\bauthor{\bsnm{Wu}, \binits{Y.N.}}:
\batitle{Deep learning with tensorflow: A review}.
\bjtitle{Journal of Educational and Behavioral Statistics}
\bvolume{45}(\bissue{2}),
\bfpage{227}--\blpage{248}
(\byear{2020})
\end{barticle}
\endbibitem

\bibitem[\protect\citeauthoryear{Imambi et~al.}{2021}]{imambi2021pytorch}
\begin{botherref}
\oauthor{\bsnm{Imambi}, \binits{S.}},
\oauthor{\bsnm{Prakash}, \binits{K.B.}},
\oauthor{\bsnm{Kanagachidambaresan}, \binits{G.}}:
Pytorch.
Programming with TensorFlow: solution for edge computing applications,
87--104
(2021)
\end{botherref}
\endbibitem

\end{thebibliography}

\end{document}